%% file: acl2021.tex
\documentclass[11pt,a4paper]{article}
\usepackage[hyperref]{acl2021}
\usepackage{times}
\usepackage{latexsym}

\usepackage{microtype}

\renewenvironment{itemize}[1]{\begin{compactitem}#1}{\end{compactitem}}
\renewenvironment{enumerate}[1]{\begin{compactenum}#1}{\end{compactenum}}

\usepackage{amssymb}% http://ctan.org/pkg/amssymb
\usepackage{pifont}% http://ctan.org/pkg/pifont
\newcommand{\cmark}{\ding{51}}%
\newcommand{\xmark}{\ding{55}}%

\usepackage{pbox}
\usepackage{hyperref}
\usepackage{url}
\usepackage{comment}
\usepackage{subfig}
\usepackage[flushleft]{threeparttable}
\usepackage{graphicx}
\usepackage{amsmath}
\usepackage{tabularx}
\usepackage{multirow, makecell}
\usepackage{algorithm}
\usepackage{xcolor}
\usepackage{booktabs}
\usepackage{paralist}
\DeclareMathOperator*{\softmax}{softmax}

\usepackage[multiple]{footmisc}
\usepackage{progressbar}
\usepackage[normalem]{ulem}

\usepackage{cleveref}
\crefname{section}{§}{§§}
\Crefname{section}{§}{§§}

\usepackage{arydshln}
\usepackage{fdsymbol}
\usepackage[utf8]{inputenc}
\usepackage{cleveref}
\crefname{section}{§}{§§}
\Crefname{section}{§}{§§}

\renewenvironment{enumerate}[1]{\begin{compactenum}#1}{\end{compactenum}}

\newcommand{\dataset}{\textsf{PolicyIE}}

\definecolor{green}{rgb}{0.0, 0.5, 0.0}
\definecolor{chocolate}{RGB}{210,105,30}
\definecolor{msblue}{RGB}{123, 104, 238}
\definecolor{orchid}{RGB}{218,112,214}
\definecolor{orchid}{RGB}{218,112,214}
\definecolor{brown}{RGB}{139,69,19}

\aclfinalcopy % Uncomment this line for the final submission
\def\aclpaperid{1010} %  Enter the acl Paper ID here

\title{Intent Classification and Slot Filling for Privacy Policies}

\author{
Wasi Uddin Ahmad$^\dagger$\thanks{~~Equal contribution. Listed by  alphabetical order.}, Jianfeng Chi$^\ddagger$\footnotemark[1], Tu Le$^\ddagger$ \\
\textbf{Thomas Norton}$^\S$, \textbf{Yuan Tian}$^\ddagger$, \textbf{Kai-Wei Chang}$^\dagger$ \\
$^\dagger$University of California, Los Angeles, $^\ddagger$University of Virginia, $^\S$Fordham University \\
\texttt{$^\dagger${\{wasiahmad, kwchang\}@cs.ucla.edu}}\\  \texttt{$^\ddagger${\{jc6ub,tnl6wk,yuant\}@virginia.edu}}\\
\texttt{$^\S${tnorton1@law.fordham.edu}}
}

\date{}

\begin{document}

\setlength{\abovedisplayskip}{4pt}
\setlength{\belowdisplayskip}{4pt}

\maketitle

\begin{abstract}

% However, it is challenging for users to read and comprehend policy documents as they can be thousands of words long.
Understanding privacy policies is crucial for users as it empowers them to learn about the information that matters to them.
Sentences written in a privacy policy document explain privacy practices, and the constituent text spans convey further specific information about that practice.
We refer to predicting the privacy practice explained in a sentence as intent classification and identifying the text spans sharing specific information as slot filling.
In this work, we propose {\dataset}, an English corpus consisting of 5,250 intent and 11,788 slot annotations spanning 31 privacy policies of websites and mobile applications.
{\dataset} corpus is a challenging real-world benchmark with limited labeled examples reflecting the cost of collecting large-scale annotations from domain experts.
We present two alternative neural approaches as baselines, (1) intent classification and slot filling as a joint sequence tagging and (2) modeling them as a sequence-to-sequence (Seq2Seq) learning task.
The experiment results show that both approaches perform comparably in intent classification, while the Seq2Seq method outperforms the sequence tagging approach in slot filling by a large margin.
We perform a detailed error analysis to reveal the challenges of the proposed corpus.
% Error analysis reveals the deficiency of the baseline approaches, suggesting room for improvement in future works.
% We hope the {\dataset} corpus will stimulate future research in this domain.
\end{abstract}

\input{sections/intro}
\input{sections/data}

\input{sections/method}
\input{sections/experiment}
\input{sections/relwork}

\input{sections/conclusion}

\section*{Acknowledgments}
The authors acknowledge the law students Michael Rasmussen and Martyna Glaz at Fordham University who worked as annotators to make the development of this corpus possible. This work was supported in part by National Science Foundation Grant OAC 1920462. Any opinions, findings, conclusions, or recommendations expressed herein are those of the authors, and do not necessarily reflect those of the US Government or NSF.

% \clearpage
\section*{Broader Impact}
Privacy and data breaches have a significant impact on individuals.
In general, security breaches expose the users to different risks such as financial loss (due to losing employment or business opportunities), physical risks to safety, and identity theft. 
Identity theft is among the most severe and fastest-growing crimes.
However, the risks due to data breaches can be minimized if the users know their rights and how they can exercise them to protect their privacy.
This requires the users to read the privacy policies of websites they visit or the mobile applications they use.
As reading privacy policies is a tedious task, automating privacy policy analysis reduces the burden of users.
Automating information extraction from privacy policies empowers users to be aware of their data collected and analyzed by service providers for different purposes.
Service providers collect consumer data at a massive scale and often fail to protect them, resulting in data breaches that have led to increased attention towards data privacy and related risks.
Reading privacy policies to understand users' rights can help take informed and timely decisions on safeguarding data privacy to mitigate the risks.
Developing an automated solution to facilitate policy document analysis requires labeled examples, and the {\dataset} corpus adds a new dimension to the available datasets in the security and privacy domain.
While {\dataset} enables us to train models to extract fine-grained information from privacy policies, the corpus can be coupled with other existing benchmarks to build a comprehensive system.
For example, PrivacyQA corpus~\citep{Ravichander2019Question} combined with {\dataset} can facilitate building QA systems that can answer questions with fine-grained details.
We believe our experiments and analysis will help direct future research.

\bibliography{anthology,acl2021}
\bibliographystyle{acl_natbib}

\clearpage
\appendix

\input{appendix}

\end{document}

%% file: sections/intro.tex
\section{Introduction}

Privacy policies inform users about how a service provider collects, uses, and maintains the users' information. 
% Privacy Policies specify how companies collect, share, maintain, and protect users' data via websites and mobile applications.
The service providers collect the users' data via their websites or mobile applications and analyze them for various purposes.
The users' data often contain sensitive information; therefore, the users must know how their information will be used, maintained, and protected from unauthorized and unlawful use.
Privacy policies are meant to explain all these use cases in detail.
This makes privacy policies often very long, complicated, and confusing~\citep{mcdonald2008cost, reidenberg2016ambiguity}.
As a result, users do not tend to read privacy policies \citep{federal2012protecting, gluck2016short, marotta2015does}, leading to undesirable consequences.
% The disconnect between users' actual behavior and hard-to-understand privacy policies could lead to some undesirable consequences. 
For example, users might not be aware of their data being sold to third-party advertisers even if they have given their consent to the service providers to use their services in return.
Therefore, automating information extraction from verbose privacy policies can help users understand their rights and make informed decisions.

% Before visiting the website and using the mobile applications, user need to know about the privacy practices (e.g., with whom the company shares users' data or how the company protects the users' data) that are pertinent to them and then decide to whether they should accept or decline those practices. However, privacy policies tend to be long and hard to understand~\citep{mcdonald2008cost, reidenberg2016ambiguity} and prior works show that people generally do not read them~\cite{federal2012protecting, gluck2016short, marotta2015does}. The disconnect between users' actual behavior and hard-to-understand privacy policies could lead to some undesirable consequences. For example, users might not be aware of their data to be sold to third party advertisers even if they do not agree to do so. 

In recent years, we have seen substantial efforts to utilize natural language processing (NLP) techniques to automate privacy policy analysis.
In literature, information extraction from policy documents is formulated as text classification \citep{wilson2016creation, harkous2018polisis, zimmeck2019maps}, text alignment \citep{liu2014step,ramanath2014unsupervised}, and question answering (QA) \citep{shvartzshanider2018recipe, harkous2018polisis, Ravichander2019Question, ahmad-etal-2020-policyqa}.
Although these approaches effectively identify the sentences or segments in a policy document relevant to a privacy practice, they lack in extracting fine-grained structured information.
As shown in the first example in Table \ref{table:example}, the privacy practice label ``Data Collection/Usage'' informs the user how, why, and what types of user information will be collected by the service provider.
The policy also specifies that users' ``username'' and ``icon or profile photo'' will be used for ``marketing purposes''.
This informs the user precisely what and why the service provider will use users' information.
% , allowing them to make an informed decision.
% However, most prior works frame information extraction by utilizing only the underlying privacy practice label for sentences or segments to identify relevant information for users.
% For example, \citet{harkous2018polisis} considered if a user query and a policy segment has the same class label, they are relevant.
% \citet{Ravichander2019Question} framed QA as identifying all the relevant sentences in a policy document for the user query.

\input{tables/ie_examples}

% As an example, \citet{Ravichander2019Question} proposed a corpus of 35 policy documents with an average length $\sim$3000 words and binary relevance labels between sentences and questions collected from crowdworkers.
% Therefore, prior works annotated privacy policies at the sentence level, without further zooming into the constituent text spans that convey specific information.
% In this work, we frame information extraction from privacy policies as intent classification and slot filling.
The challenge in training models to extract fine-grained information is the lack of labeled examples.
Annotating privacy policy documents is expensive as they can be thousands of words long and requires domain experts (e.g., law students).
Therefore, prior works annotate privacy policies at the sentence level, without further utilizing the constituent text spans to convey specific information.
Sentences written in a policy document explain privacy practices, which we refer to as \emph{intent classification} and identifying the constituent text spans that share further specific information as \emph{slot filling}. 
Table \ref{table:example} shows a couple of examples.
This formulation of information extraction lifts users' burden to comprehend relevant segments in a policy document and identify the details, such as how and why users' data are collected and shared with others.

To facilitate fine-grained information extraction, we present {\dataset}, an English corpus consisting of 5,250 intent and 11,788 slot annotations over 31 privacy policies of websites and mobile applications.
We perform experiments using sequence tagging and sequence-to-sequence (Seq2Seq) learning models to jointly model intent classification and slot filling.
The results show that both modeling approaches perform comparably in intent classification, while Seq2Seq models outperform the sequence tagging models in slot filling by a large margin.
We conduct a thorough error analysis and categorize the errors into seven types.
We observe that sequence tagging approaches miss more slots while Seq2Seq models predict more spurious slots.
We further discuss the error cases by considering other factors to help guide future work.
% (e.g., slot length and their frequency in training examples) 
We release the code and data to facilitate research.\footnote{https://github.com/wasiahmad/PolicyIE}

%% file: tables/ie_examples.tex
%%%%%%%%%%%%%%%%%%%%%%%%%%%%%%%%%%%%%%%%%%%%%%%%%%%%%%%%%%%%%%%%
\begin{table*}[t]
\centering
% \vspace{4mm}
% \small
\def\arraystretch{1.2}%
\resizebox{0.98\linewidth}{!}{%
\begin{tabular}{p{0.99\linewidth}} % p{0.99\linewidth}
\hline
% gg \\
% \hline
\textcolor{orange}{$\text{[We]}_{\text{Data Collector: First Party Entity}}$} may also \textcolor{green}{$\text{[use]}_{\text{Action}}$} or display \textcolor{purple}{$\text{[your]}_{\text{Data Provider: user}}$}\\\textcolor{blue}{$\text{[username]}_{\text{Data Collected: User Online Activities/Profiles}}$} and \textcolor{blue}{$\text{[icon or profile photo]}_{\text{Data Collected: User Online Activities/Profiles}}$}\\on \textcolor{magenta}{$\text{[marketing purpose or press releases]}_{\text{Purpose: Advertising/Marketing}}$}.
\\
\multicolumn{1}{l}{\emph{Privacy Practice}. Data Collection/Usage}
\\
\hline
\textcolor{chocolate}{$\text{[We]}_{\text{Data Sharer: First Party Entity}}$} do \textcolor{brown}{$\text{[not]}_{\text{Polarity: Negation}}$} \textcolor{green}{$\text{[sell]}_{\text{Action}}$} \textcolor{purple}{$\text{[your]}_{\text{Data Provider: user}}$}\\\textcolor{msblue}{$\text{[personal information]}_{\text{Data Shared: General Data}}$} to \textcolor{cyan}{$\text{[third parties]}_{\text{Data Receiver: Third Party Entity}}$}.
\\
\multicolumn{1}{l}{\emph{Privacy Practice}. Data Sharing/Disclosure}
\\
\hline
\end{tabular}
}
\vspace{-1mm}
\caption{Annotation examples from \dataset~Corpus. Best viewed in color.}
\vspace{-2mm}
\label{table:example}
\end{table*}

%% file: sections/data.tex
\section{Construction of PolicyIE Corpus}

% In this section, we describe the methodology to construct the {\dataset} corpus that includes the process of selecting the policy documents for annotation (\cref{subsec:doc_selection}) followed by the details of the data annotation procedure and the format of the final corpus (\cref{sec:ann-schema}).
% , and the format of the final corpus (\cref{sec:data_format}).

\subsection{Privacy Policies Selection}  
\label{subsec:doc_selection}
% Privacy policies are lengthy and section-based documents which describe how the service provider collects, shares, stores and secures users' data in its websites or apps.
% In general, privacy policy documents are organized in sections that describe how service providers collect, share, store, and protect consumer data.
The scope of privacy policies primarily depends on how service providers function.
For example, service providers primarily relying on mobile applications (e.g., Viber, Whatsapp) or websites and applications (e.g., Amazon, Walmart) have different privacy practices detailed in their privacy policies.
In {\dataset}, we want to achieve broad coverage across privacy practices exercised by the service providers such that the corpus can serve a wide variety of use cases.
Therefore, we go through the following steps to select the policy documents.

\paragraph{Initial Collection}
% \smallskip
% \noindent\textbf{Initial Collection\hspace{1.5ex}}
\citet{ramanath2014unsupervised} introduced a corpus of 1,010 privacy policies of the top websites ranked on \texttt{Alexa.com}.
We crawled those websites' privacy policies in November 2019 since the released privacy policies are outdated.
For mobile application privacy policies, we scrape application information from Google Play Store using \texttt{play-scraper} public API\footnote{https://github.com/danieliu/play-scraper} and crawl their privacy policy.
We ended up with 7,500 mobile applications' privacy policies. 

% To ensure the quality of our corpus, we use the following steps to collect privacy policies.
% First, we collect privacy policies from two sources \wasi{They are not ``sources''. We should rather say, we collect privacy policies of service providers having websites or mobile applications.}: website privacy policies of mobile application privacy policies. 
% First, we collect privacy policies of service providers having websites or mobile applications.
% For the collection of website ones, we use the Privacy Policy Corpus introduced by~\citet{ramanath2014unsupervised} that contains 1,010 privacy policies from the top websites ranked on \texttt{Alexa.com}. We retrieve the latest website privacy policies from those top websites since the corpus was collected before 2014.

\paragraph{Filtering}
% \smallskip
% \noindent\textbf{Filtering\hspace{1.5ex}}
First, we filter out the privacy policies written in a non-English language and the mobile applications' privacy policies with the app review rating of less than 4.5.
Then we filter out privacy policies that are too short ($<$ 2,500 words) or too long ($>$ 6,000 words).
Finally, we randomly select 200 websites and mobile application privacy policies each (400 documents in total).\footnote{We ensure the mobile applications span different application categories on the Play Store.}
% Then, we pick the privacy policies having a moderate length (e.g., 2,500-6,000 words). 
% For app privacy policies, we further select those documents with app review scores higher than 4.5. 

\paragraph{Post-processing}
% \smallskip
% \noindent\textbf{Post-processing\hspace{1.5ex}}
We ask a domain expert (working in the security and privacy domain for more than three years) to examine the selected 400 privacy policies.
The goal for the examination is to ensure the policy documents cover the four privacy practices: (1) \emph{Data Collection/Usage}, (2) \emph{Data Sharing/Disclosure}, (3) \emph{Data Storage/Retention}, and (4) \emph{Data Security/Protection}. 
These four practices cover how a service provider processes users' data in general and are included in the General Data Protection Regulation (GDPR).
Finally, we shortlist 50 policy documents for annotation, 25 in each category (websites and mobile applications).

% The selection criterion in the step is that each document should mention all of the following privacy practices: (1) \emph{Data Collection/Usage}, (2) \emph{Data Sharing/Disclosure}, (3) \emph{Data Storage/Retention}, and (4) \emph{Data Security/Protection}. These four privacy practices describe how the service provider processes users' data in general and they are all covered by regulations of the European Union’s General Data Protection Regulation (GDPR). 
% We detail the privacy practices in Sec.~\ref{sec:ann-schema}.
% Finally, we select 31 out of 400 privacy policies (16 from websites and 15 from mobile applications).
% \wasi{Overall, the writing of section 2.1 looks a bit unorganized.}

\subsection{Data Annotation}  
\label{sec:ann-schema}

% \paragraph{Annotation Schema}
% We formulate our annotation as privacy practices prediction with slot filling. 
% Specifically, we consider privacy practices mentioned above: 
% Specifically, we categorize private policy sentences under five categories and consider the first four categories from the annotation schema suggested by~\citet{wilson2016creation}.

\paragraph{Annotation Schema}
% \smallskip
% \noindent\textbf{Annotation Schema\hspace{1.5ex}}
To annotate sentences in a policy document, we consider the first four privacy practices from the annotation schema suggested by~\citet{wilson2016creation}.
Therefore, we perform sentence categorization under five \emph{intent classes} that are described below.
\begin{enumerate}
    \setlength{\itemindent}{0em}
    \item[(1)] \emph{Data Collection/Usage}: What, why and how user information is collected;
    \item[(2)] \emph{Data Sharing/Disclosure}: What, why and how user information is shared with or collected by third parties; 
    \item[(3)] \emph{Data Storage/Retention}: How long and where user information will be stored;
    \item[(4)] \emph{Data Security/Protection}: Protection measures for user information;
    \item[(5)] \emph{Other}: Other privacy practices that do not fall into the above four categories.
\end{enumerate}

Apart from annotating sentences with privacy practices, we aim to identify the text spans in sentences that explain specific details about the practices.
For example, in the sentence \emph{``we collect personal information in order to \uline{provide users with a personalized experience}''}, the underlined text span conveys the purpose of data collection.
In our annotation schema, we refer to the identification of such text spans as \emph{slot filling}.
There are 18 slot labels in our annotation schema (provided in Appendix).
% as listed in Table \ref{table:slot_types}.
We group the slots into two categories: type-I and type-II based on their role in privacy practices.
% as well as their differences in our annotations (e.g., frequency and length).
While the type-I slots include participants of privacy practices, such as \emph{Data Provider}, \emph{Data Receiver}, type-II slots include purposes, conditions that characterize more details of privacy practices. 
Note that type-I and type-II slots may overlap, e.g., in the previous example, the underlined text span is the \emph{purpose} of data collection, and the span ``user'' is the \emph{Data Provider} (whose data is collected).
In general, type-II slots are longer (consisting of more words) and less frequent than type-I slots.

In total, there are 14 type-I and 4 type-II slots in our annotation schema.
These slots are associated with a list of attributes, e.g., \emph{Data Collected} and \emph{Data Shared} have the attributes \emph{Contact Data}, \emph{Location Data}, \emph{Demographic Data}, etc.
Table~\ref{table:example} illustrates a couple of examples. 
We detail the slots and their attributes in the Appendix.
\input{tables/data_stats}

\input{tables/model_input_output}

\paragraph{Annotation Procedure}
% \smallskip
% \noindent\textbf{Annotation Procedure\hspace{1.5ex}}
General crowdworkers such as Amazon Mechanical Turkers are not suitable to annotate policy documents as it requires specialized domain knowledge \cite{mcdonald2008cost, reidenberg2016ambiguity}.
We hire two law students to perform the annotation.
% We describe in detail how we annotate privacy policies following our annotation schema. 
% \wasi{``privacy policies'' in place of ``our corpus''}
% to annotate the documents since privacy policies are difficult for normal human workers such as Amazon Mechanical Turkers to comprehend~\citep{mcdonald2008cost,reidenberg2016ambiguity}.  
We use the web-based annotation tool, \textsc{BRAT}~\citep{stenetorp2012brat} to conduct the annotation. 
We write a detailed annotation guideline and pretest them through multiple rounds of pilot studies.
The guideline is further updated with notes to resolve complex or corner cases during the annotation process.\footnote{We release the guideline as supplementary material.}
The annotation process is closely monitored by a domain expert and a legal scholar 
and is granted IRB exempt by the Institutional Review Board (IRB).
% and it is approved by the Institutional Review Board (IRB).
The annotators are presented with one segment from a policy document at a time
% instead of the full document 
and asked to perform annotation following the guideline.
We manually segment the policy documents such that a segment discusses similar issues to reduce ambiguity at the annotator end.
The annotators worked 10 weeks, with an average of 10 hours per week, and completed annotations for 31 policy documents. 
Each annotator is paid \$15 per hour.

\paragraph{Post-editing and Quality Control}
% \smallskip
% \noindent\textbf{Post-editing and Quality Control\hspace{1.5ex}}
We compute an inter-annotator agreement for each annotated segment of policy documents using Krippendorff's Alpha ($\alpha_K$) \citep{klaus1980content}. 
% We use Krippendorff's Alpha ($\alpha_K$) \citep{klaus1980content} to measure the inter-annotator agreement. 
The annotators are asked to discuss their annotations and re-annotate those sections with token-level $\alpha_K$ falling below 0.75. 
An $\alpha_K$ value within the range of 0.67 to 0.8 is allowed for tentative conclusions \cite{artstein2008inter,reidsma2008reliability}. 
% Since type-I and type-II slots are different in terms of their roles, frequencies, and lengths, we calculate the agreement for the two categories of slots individually. 
After the re-annotation process, we calculate the agreement for the two categories of slots individually.
The inter-annotator agreement is 0.87 and 0.84 for type-I and type-II slots, respectively. 
Then the adjudicators discuss and finalize the annotations. 
The adjudication process involves one of the annotators, the legal scholar, and the domain expert.

\paragraph{Data Statistics \& Format}
% \smallskip
% \noindent\textbf{Data Statistics \& Format\hspace{1.5ex}}
Table~\ref{table:statistics} presents the statistics of the {\dataset} corpus.
The corpus consists of 15 and 16 privacy policies of websites and mobile applications, respectively.
We release the annotated policy documents split into sentences.\footnote{We split the policy documents into sentences using UDPipe~\cite{straka2016udpipe}.} 
Each sentence is associated with an intent label, and the constituent words are associated with a slot label (following the BIO tagging scheme).
% where a document consists of a list of sentences.
% 
% and every sentence is associated with 1 of the 5 intent classes, and the constituent words are associated with a slot label.
% (following the BIO tagging scheme).

%% file: tables/data_stats.tex
\begin{table}[t]
\centering
% \resizebox{\linewidth}{!}{%
% \small
% \def\arraystretch{1.1}%
% \begin{tabular}{l@{\hskip 0.05in}|@{\hskip 0.05in}c@{\hskip 0.05in}|@{\hskip 0.05in}c@{\hskip 0.05in}}
% \def\arraystretch{1.25}%
\vspace{2mm}
\begin{tabular}{l|r r}
\hline
Dataset & Train & Test  \\ 
\hline\hline
\# Policies & 25 & 6  \\
\# Sentences & 4,209 & 1,041  \\
\# Type-I slots & 7,327 & 1,704 \\
\# Type-II slots & 2,263 & 494 \\
\hline
Avg. sentence length & 23.73 & 26.62 \\
Avg. \# type-I slot / sent.  & 4.48 & 4.75 \\
Avg. \# type-II slot / sent. & 1.38 & 1.38 \\
Avg. type-I slot length & 2.01 & 2.15 \\
Avg. type-II slot length & 8.70 & 10.70 \\
\hline
\end{tabular}
% }
\vspace{-1mm}
\caption{
Statistics of the \dataset~Corpus. 
% We calculate \# of type-I(II) slots / sentence excluding the sentences with ``\emph{Other}'' intents.
}
% \vspace{-2mm}
\label{table:statistics}
\end{table}

%% file: tables/model_input_output.tex
%%%%%%%%%%%%%%%%%%%%%%%%%%%%%%%%%%%%%%%%%%%%%%%%%%%%%%%%%%%%%%%%
% \begin{table*}[ht]
% \centering
% \begin{tabular}{l} %
% \hline
% \textbf{Input:} We may also use or display your username and icon or profile photo on marketing\\ purpose or press releases .\\
% \hline
% \textbf{Joint intent and slot prediction:} \\
% Intent label: First Party Collection/Usage\\
% Type-I tags: B-DC-er.1st-P O O B-Act O O B-DPer.Usr B-DC.D-UOAP O\\
% B-DC.D-UOAP I-DC.D-UOAP I-DC.D-UOAP I-DC.D-UOAP O O O O O O O\\
% Type-II tags: O O O O O O O O O O O O O O B-Pur.AM I-Pur.AM I-Pur.AM I-Pur.AM I-Pur.AM O\\
% \hline
% \textbf{Seq2Seq model prediction:} \\
% \text{[IN: Data Collection/Usage} \text{[ARG: Data-Collector.First-Party-Entity \emph{We}]} \text{[ARG: Action \emph{use}]}\\
% \text{[ARG: Data-Provider.User \emph{your}]}\\
% \text{[ARG: Data Collected.User-Online-Activities-Profiles \emph{username}]}\\
% \text{[ARG: Data-Collected.User-Online-Activities-Profiles \emph{icon or profile photo}]}\\
% \text{[ARG: Purpose.Advertising-Marketing \emph{marketing purpose or press releases}]]}\\
% \hline
% \end{tabular}
% \caption{An Example of Input and output in our models.}
% \label{table:model-input-output}
% \end{table*}

\begin{table*}[t]
\centering
\def\arraystretch{1.1}%
\resizebox{\linewidth}{!}{%
\begin{tabular}{l} %
\hline
\textbf{Joint intent and slot tagging} \\ \hdashline
\textbf{Input:} [CLS] We may also use or display your username and icon or profile photo on marketing\\ purpose or press releases .\\ \hdashline
\textbf{Type-I slot tagging output} \\ 
Data-Collection-Usage B-DC.FPE O O B-Action O O B-DP.U B-DC.UOAP O B-DC.UOAP \\ I-DC.UOAP I-DC.UOAP I-DC.UOAP O O O O O O O \\ \hdashline
\textbf{Type-II slot tagging output} \\
Data-Collection-Usage O O O O O O O O O O O O O O B-P.AM I-P.AM I-P.AM I-P.AM I-P.AM O \\ [2pt]
\hline
\textbf{Sequence-to-sequence (Seq2Seq) learning} \\ \hdashline
\textbf{Input:} We may also use or display your username and icon or profile photo on marketing purpose \\ or press releases .\\ \hdashline
\textbf{Output:} \text{[IN:Data-Collection-Usage} \text{[SL:DC.FPE \emph{We}]} \text{[SL:Action \emph{use}]}
\text{[SL:DP.U \emph{your}]} 
\text{[SL:DC.UOAP} \\ \text{\emph{username}]}
\text{[SL:DC.UOAP \emph{icon or profile photo}]}
\text{[SL:P.AM \emph{marketing purpose or press releases}]]} \\ [2pt]
% \hline 
% Symbols used in labels \\
% DP.U = Data-Provider.User ;
% DC.FPE = Data-Collector.First-Party-Entity \\
% DC.UOAP = Data-Collected.User-Online-Activities-Profiles ;
% P.AM = Purpose.Advertising-Marketing \\
\hline
\end{tabular}
}
\vspace{-1mm}
\caption{
An example of input / output used to train the two types of models on {\dataset}.
For brevity, we replaced part of label strings with symbols: DP.U, DC.FPE, DC.UOAP, P.AM represents Data-Provider.User, Data-Collector.First-Party-Entity, Data-Collected.User-Online-Activities-Profiles, and Purpose.Advertising-Marketing.
}
% \vspace{-2mm}
\label{table:model-input-output}
\end{table*}

%% file: sections/method.tex
\section{Model \& Setup}
\label{sec:model}

{\dataset} provides annotations of privacy practices and corresponding text spans in privacy policies.
We refer to privacy practice prediction for a sentence as \emph{intent classification} and identifying the text spans as \emph{slot filling}.
% We want to utilize our dataset to train models that can identify the intent and corresponding slots behind each sentence in a privacy policy.
We present two alternative approaches; the first approach jointly models intent classification and slot tagging \cite{chen2019bert}, and the second modeling approach casts the problem as a sequence-to-sequence learning task \cite{rongali2020don, li2020mtop}.

\input{tables/seqtag}

\subsection{Sequence Tagging}
Following \citet{chen2019bert}, given a sentence $s = w_1, \ldots, w_l$ from a privacy policy document $D$, a special token ($ w_0 = [\text{CLS}]$) is prepended to form the input sequence that is fed to an encoder.
The encoder produces contextual representations of the input tokens $h_0, h_1, \ldots, h_l$ where $h_0$ and $h_1, \ldots, h_l$ are fed to separate $\softmax$ classifiers to predict the target intent and slot labels.
\begin{align*}
    y^i =& \softmax(W^T_i h_0 + b_i), \\
    y^s_n =& \softmax(W^T_s h_n + b_s), n \in 1, \ldots l,
\end{align*}
where $W_i \in R^{d \times I}, W_s \in R^{d \times S}, b_r \in R^I$ and $b_i \in R^I, b_s \in R^S$ are parameters, and $I, S$ are the total number of intent and slot types, respectively.
% To jointly model intent classification and slot filling, the model learns to maximize the conditional probability:
The sequence tagging model (composed of an encoder and a classifier) learns to maximize the following conditional probability to perform intent classification and slot filling jointly.
\begin{equation*}
    P(y^i, y^s | s) = p(y^i|s) \prod_{n=1}^{l} p(y^s_n|s).
\end{equation*}

We train the models end-to-end by minimizing the cross-entropy loss.
Table \ref{table:model-input-output} shows an example of input and output to train the joint intent and slot tagging models.
Since type-I and type-II slots have different characteristics as discussed in \cref{sec:ann-schema} and overlap, we train two separate sequential tagging models for type-I and type-II slots to keep the baseline models simple.\footnote{Span enumeration based techniques \cite{wadden-etal-2019-entity, luan-etal-2019-general} can be utilized to perform tagging both types of slots jointly, and we leave this as future work.}
% We individually train sequence tagging models for type-I and type-II slots as they overlap.
We use BiLSTM~\cite{liu2016attention, zhang2016joint}, Transformer~\cite{vaswani2017attention}, BERT~\cite{vaswani2017attention}, and RoBERTa~\cite{liu2019roberta} as encoder to form the sequence tagging models.

% \subsubsection{Encoders}
% \smallskip
% \noindent$\bullet$~\textbf{Feature}
% \smallskip
% \noindent\textbf{Embedding \hspace{0.5em}} 

Besides, we consider an embedding based baseline where the input word embeddings are fed to the $\softmax$ classifiers. The special token ($ w_0 = [\text{CLS}]$) embedding is formed by applying average pooling over the input word embeddings.
We train WordPiece embeddings with a 30,000 token vocabulary \cite{devlin2019bert} using \emph{fastText} \cite{bojanowski2017enriching} based on a corpus of 130,000 privacy policies collected from apps on the Google Play Store \cite{harkous2018polisis}.
We use the hidden state corresponding to the first WordPiece of a token to predict the target slot labels.
% To make slot label prediction compatible with WordPieces, we use the hidden state corresponding to the first WordPiece of a token as input to the $\softmax$ classifier.

% \smallskip
% \noindent\textbf{BiLSTM \hspace{0.5em}}~\cite{liu2016attention, zhang2016joint}
% \smallskip
% \noindent\textbf{Transformer \hspace{0.5em}}~\cite{vaswani2017attention} 
% \smallskip
% \noindent\textbf{BERT \hspace{0.5em}}~\cite{devlin2019bert} 
% \smallskip
% \noindent\textbf{RoBERTa \hspace{0.5em}}~\cite{liu2019roberta} 

\paragraph{Conditional Random Field (CRF)}
% \smallskip
% \noindent\textbf{Conditional Random Field (CRF)\hspace{1.5ex}} 
helps structure prediction tasks, such as semantic role labeling \cite{zhou-xu-2015-end} and named entity recognition \cite{cotterell-duh-2017-low}.
% Therefore, we investigate the effectiveness of applying CRF in capturing slot label dependencies following \citet{chen2019bert}.
Therefore, we model slot labeling jointly using a conditional random
field (CRF) \cite{lafferty2001conditional} (only interactions between two successive labels are considered).
We refer the readers to \citet{ma-hovy-2016-end} for details.
% Training and decoding can be solved efficiently by adopting the Viterbi algorithm.
% We use the \emph{forward-backward} algorithm to compute the log-likelihood of the inputs assuming a CRF model.

\subsection{Sequence-to-Sequence Learning}
Recent works in semantic parsing~\citep{rongali2020don, zhu-etal-2020-dont, li2020mtop} formulate the task as sequence-to-sequence (Seq2Seq) learning.
Taking this as a motivation, we investigate the scope of Seq2Seq learning for joint intent classification and slot filling for privacy policy sentences.
In Table \ref{table:model-input-output}, we show an example of encoder input and decoder output used in Seq2Seq learning.
We form the target sequences by following the template: [IN:LABEL [SL:LABEL $w_1, \ldots, w_m$] \ldots ].
% by concatenating the intent and slot labels and follow a template.
During inference, we use greedy decoding and parse the decoded sequence to extract intent and slot labels.
Note that we only consider text spans in the decoded sequences that are surrounded by ``[]''; the rest are discarded.
Since our proposed {\dataset} corpus consists of a few thousand examples, instead of training Seq2Seq models from scratch, we fine-tune pre-trained models as the baselines.
Specifically, we consider five state-of-the-art models: MiniLM~\cite{wang2020minilm}, UniLM~\cite{dong2019unified}, UniLMv2~\cite{bao2020unilmv2}, MASS~\cite{song2019mass}, and BART~\cite{lewis-etal-2020-bart}.

\input{tables/seq2seq}

\subsection{Setup}
\label{sec:setup}
\paragraph{Implementation}
% \smallskip
% \noindent\textbf{Implementation\hspace{1.5ex}}
We use the implementation of BERT and RoBERTa from \texttt{transformers} API \cite{wolf-etal-2020-transformers}.
For the Seq2Seq learning baselines, we use their public implementations.\footnote{\url{https://github.com/microsoft/unilm}}\footnote{\url{https://github.com/microsoft/MASS}}\footnote{\url{https://github.com/pytorch/fairseq/tree/master/examples/bart}}
We \emph{train} BiLSTM, Transformer baseline models and \emph{fine-tune} all the other
% BERT, RoBERTa, MiniLM, UniLM, UniLMv2, MASS, and BART
baselines for 20 epochs and choose the best checkpoint based on validation performance.
From 4,209 training examples, we use 4,000 examples for training ($\sim$95\%) and 209 examples for validation ($\sim$5\%).
% \footnote{From 4,209 training examples, we use 4,000 examples for training ($\sim$95\%) and 209 examples for validation ($\sim$5\%).}
We tune the learning rate in [1e-3, 5e-4, 1e-4, 5e-5, 1e-5] and set the batch size to 16 in all our experiments (to fit in one GeForce GTX 1080 GPU with 11gb memory).
We train (or fine-tune) all the models five times with different seeds and report average performances.

\paragraph{Evaluation Metrics}
% \smallskip
% \noindent\textbf{Evaluation Metrics\hspace{1.5ex}}
To evaluate the baseline approaches, we compute the F1 score for intent classification and slot filling tasks.\footnote{We use a micro average for intent classification.}
We also compute an exact match (EM) accuracy (if the predicted intent matches the reference intent and slot F1 = 1.0).

\paragraph{Human Performance}
% \smallskip
% \noindent\textbf{Human Performance\hspace{1.5ex}}
is computed by considering each annotator's annotations as predictions and the adjudicated annotations as the reference. The final score is an average across all annotators.

%% file: tables/seqtag.tex
\begin{table*}[!ht]
\centering
% \resizebox{\linewidth}{!}{%
% \small
% \def\arraystretch{1.1}%
\begin{tabular}{l|c|c|c c|c c}
\hline
\multirow{2}{*}{Model} & \multirow{1}{*}{\# param} & \multirow{2}{*}{Intent F1} & \multicolumn{2}{c|}{Type-I} & \multicolumn{2}{c}{Type-II} \\ 
\cline{4-7}
& (in millions) & & \multicolumn{1}{c}{Slot F1} & \multicolumn{1}{c|}{EM} & \multicolumn{1}{c}{Slot F1} & \multicolumn{1}{c}{EM} \\ 
\hline
% Human & - & 96.5$_{\pm 0.5}$ & 84.3$_{\pm 0.6}$ & 56.6$_{\pm 2.8}$ & 62.3$_{\pm 3.5}$ & 55.6$_{\pm 4.1}$\\
Human & - & 96.5 & 84.3 & 56.6 & 62.3 & 55.6 \\
\hline
Embedding & 1.7 & 50.9$_{\pm 27.3}$ & 19.1$_{\pm 0.3}$ & 0.8$_{\pm 0.3}$ & 0.0$_{\pm 0.0}$ & 0.0$_{\pm 0.0}$ \\
BiLSTM & 8 & 75.9$_{\pm 1.1}$ & 40.8$_{\pm 0.9}$ & 7.6$_{\pm 0.9}$ & 3.9$_{\pm 3.0}$ & 10.0$_{\pm 2.7}$ \\
Transformer & 34.8 & 80.1$_{\pm 0.6}$ & 41.0$_{\pm 3.5}$ & 6.5$_{\pm 2.8}$ & 3.5$_{\pm 1.0}$ & 13.1$_{\pm 2.4}$ \\
BERT & 110 & \textbf{84.7$_{\pm 0.7}$} & 55.5$_{\pm 1.1}$ & 17.0$_{\pm 1.1}$ & 29.6$_{\pm 2.4}$ & 24.2$_{\pm 4.2}$ \\
RoBERTa & 124 & 84.5$_{\pm 0.7}$ & 54.2$_{\pm 1.9}$ & 14.3$_{\pm 2.4}$ & 29.8$_{\pm 1.7}$ & 24.8$_{\pm 1.4}$ \\ \hline
Embedding w/ CRF & 1.7 & 67.9$_{\pm 0.6}$  & 26.0$_{\pm 1.5}$ & 1.20$_{\pm 0.3}$ & 5.7$_{\pm 4.6}$ & 3.1$_{\pm 0.6}$ \\
BiLSTM w/ CRF & 8 & 76.7$_{\pm 1.4}$ & 45.1$_{\pm 1.2}$ & 9.2$_{\pm 0.9}$ & 26.8$_{\pm 2.2}$ & 18.1$_{\pm 2.0}$ \\
Transformer w/ CRF & 34.8 & 77.9$_{\pm 2.7}$ & 43.7$_{\pm 2.3}$ & 8.9$_{\pm 3.0}$ & 5.7$_{\pm 0.9}$ & 11.0$_{\pm 2.1}$ \\
BERT w/ CRF & 110 & 82.1$_{\pm 2.0}$ & 56.0$_{\pm 0.8}$ & \textbf{19.2$_{\pm 1.1}$} & 31.7$_{\pm 1.9}$ & 19.7$_{\pm 2.6}$ \\
RoBERTa w/ CRF & 124 & 83.3$_{\pm 1.6}$ & \textbf{57.0$_{\pm 0.6}$} & 18.2$_{\pm 1.2}$ & \textbf{34.5$_{\pm 1.3}$} & \textbf{27.7$_{\pm 3.9}$} \\
\hline
\end{tabular}
% }
\vspace{-1mm}
\caption{
Test set performance of the sequence tagging models on {\dataset} corpus.
% $^{\ast}$ indicates the model uses conditional random field (CRF) in slot tagging.
We individually train and evaluate the models on intent classification and type-I and type-II slots tagging and report average intent F1 score.
% $|$M$|$ indicates the number of model parameters (in millions).
% We trained all the models five times with different seeds and reporting average performances.
}
% \vspace{-1mm}
\label{table:seqtag}
\end{table*}

%% file: tables/seq2seq.tex
\begin{table*}[!ht]
\centering
% \resizebox{\linewidth}{!}{%
% \small
% \def\arraystretch{1.1}%
\begin{tabular}{l|c|c|c c|c c}
\hline
\multirow{2}{*}{Model} & \multirow{1}{*}{\# param}  & \multirow{2}{*}{Intent F1} & \multicolumn{2}{c|}{Type-I} & \multicolumn{2}{c}{Type-II} \\ 
\cline{4-7}
& (in millions) &  & Slot F1 & EM & Slot F1 & EM \\
\hline
% Human Performance & N/A & 96.5$_{\pm 0.5}$ & 83.8$_{\pm 0.6}$ & 57.1$_{\pm 3.2}$  & 67.8$_{\pm  3.1}$  & 55.4$_{\pm 4.3}$ \\
% Human & - & 96.5$_{\pm 0.5}$ & 84.3$_{\pm 0.6}$ & 56.6$_{\pm 2.8}$ & 62.3$_{\pm 3.5}$ & 55.6$_{\pm 4.1}$\\
Human & - & 96.5 & 84.3 & 56.6 & 62.3 & 55.6 \\
\hline
MiniLM & 33 & 83.9$_{\pm 0.3}$ & 52.4$_{\pm 1.5}$ & 19.8$_{\pm 1.6}$ & 40.4$_{\pm 0.4}$ & 27.9$_{\pm 1.6}$ \\
UniLM & 110 & 83.6$_{\pm 0.5}$ & 58.2$_{\pm 0.7}$ & 28.6$_{\pm 1.2}$ & 53.5$_{\pm 1.4}$ & \textbf{35.4$_{\pm 1.9}$} \\
UniLMv2 & 110 & \textbf{84.7$_{\pm 0.5}$} & 61.4$_{\pm 0.9}$ & \textbf{29.9$_{\pm 1.2}$} & 53.5$_{\pm 1.5}$ & 33.5$_{\pm 1.5}$ \\
MASS & 123 & 81.8$_{\pm 1.2}$ & 54.1$_{\pm 2.5}$ & 21.3$_{\pm 2.0}$ & 44.9$_{\pm 1.2}$ & 25.3$_{\pm 1.3}$ \\
\hline
\multirow{2}{*}{BART} & 140 & 83.3$_{\pm 1.1}$ & 53.6$_{\pm 1.7}$ & 10.6$_{\pm 1.7}$ & 52.4$_{\pm 2.7}$ & 27.5$_{\pm 2.2}$ \\
& 400 & 83.6$_{\pm 1.3}$ & \textbf{63.7$_{\pm 1.3}$} & 23.0$_{\pm 1.3}$ & \textbf{55.2$_{\pm 1.0}$} & 31.6$_{\pm 2.0}$ \\
\hline
\end{tabular}
% }
\vspace{-1mm}
\caption{
Test set performance of the Seq2Seq models on {\dataset} corpus.
% $|$M$|$ indicates the number of model parameters (in millions).
% We trained all the models five times with different seeds and reporting average performances.
}
% }
\label{table:seq2seq}
% \vspace{-2mm}
\end{table*}

%% file: sections/experiment.tex
\section{Experiment Results \& Analysis}
We aim to address the following questions.
\begin{enumerate}
\setlength{\itemindent}{0em}
\item How do the two modeling approaches perform on our proposed dataset (\cref{sec:main_result})?
\item How do they perform on different intent and slot types (\cref{sec:perf_break})?
\item What type of errors do the best performing models make (\cref{sec:error_anal})?
% Which modeling approach performs better and why? 
\end{enumerate}

\subsection{Main Results}
\label{sec:main_result}

\paragraph{Sequence Tagging} 
% \smallskip
% \noindent\textbf{Sequence Tagging\hspace{1.5ex}}
The overall performances of the sequence tagging models are presented in Table \ref{table:seqtag}.
The pre-trained models, BERT and RoBERTa, outperform other baselines by a large margin.
% We can see that BERT and RoBERTa and their CRF variants outperform the other models in terms of intent classification and slot tagging. 
% Using conditional random field (CRF), the models boost the slot tagging accuracy with a slight drop in intent classification accuracy.
Using conditional random field (CRF), the models boost the slot tagging performance with a slight degradation in intent classification performance.
For example, RoBERTa + CRF model improves over RoBERTa by 2.8\% and 3.9\% in terms of type-I slot F1 and EM with a 0.5\% drop in intent F1 score.
% outperforms its linear variant by 2.8\% and 3.9\% in terms of type-I slot F1 and type-I EM (4.7\% and 2.9\% in terms of type-II slot F1 and type-II EM) with a 0.5\% decrease in terms of intent F1. 
The results indicate that predicting type-II slots is difficult compared to type-I slots as they differ in length (type-I slots are mostly phrases, while type-II slots are clauses) and are less frequent in the training examples.
% The results also indicate that type-I slot tagging is easier than type-II slot tagging, since the slot F1 scores of type-I slot tagging are higher than those of type-II slot tagging. 
However, the EM accuracy for type-I slots is lower than type-II slots due to more type-I slots ($\sim$4.75) than type-II slots ($\sim$1.38) on average per sentence.
Note that if models fail to predict one of the slots, EM will be zero. 
% However, EMs of type-I slots are lower than those of type-I slots. 
% This is because the average number of type-I slots ~($\sim$4.75) is greater than type-II slots ~($\sim$1.38) per sentence, and if models miss one of the slots, EM will be zero. 
% on average, we have 4.75 type-I slots per sentence and 1.38 type-II slots per sentence on test set excluding the sentences with ``Other'' intent (see Table~\ref{table:statistics}). 
% If a model misses to predict one of the slots in a sentence, EM will be zero. 
% It also explains why the performance gaps between slot F1 and EM of type-I slots are obviously larger than those of type-I slots.

% Key points of Table~\ref{table:seqtag}
% (1) BERT and RoBERTa outperform the other model in terms of intent F1, slot F1 and EM.
% (2) CRF model helps to increase slot F1 and EM
% (3) slot F1: type-I > type-II; EM type-I < type-II.

\paragraph{Seq2Seq Learning} 
% \smallskip
% \noindent\textbf{Seq2Seq Learning\hspace{1.5ex}}
Seq2Seq models predict the intent and slots by generating the labels and spans following a template.
Then we extract the intent and slot labels from the generated sequences.
The experiment results are presented in Table~\ref{table:seq2seq}. 
To our surprise, we observe that all the models perform well in predicting intent and slot labels.
The best performing model is BART (according to slot F1 score) with 400 million parameters, outperforming its smaller variant by 10.1\% and 2.8\% in terms of slot F1 for type-I and type-II slots, respectively.
% Overall, all Seq2Seq models do well in intent classification with all intent F1 scores greater than 80\%. 
% The results also indicate that the performance improves with increased model size for BART-based models.

%  Key points of Table~\ref{table:seq2seq}
% (1) All models are good in terms of intent F1
% (2) With the increase of parameters in the model, the overall model performance increases. 

% \input{figures/s2s_confusion_matrix}

\paragraph{Sequence Tagging vs. Seq2Seq Learning}
% \smallskip
% \noindent\textbf{Sequence Tagging vs. Seq2Seq Learning\hspace{1.5ex}}
It is evident from the experiment results that Seq2Seq models outperform the sequence tagging models in slot filling by a large margin, while in intent classification, they are competitive.
However, both the modeling approaches perform poorly in predicting all the slots in a sentence correctly, resulting in a lower EM score.
% The Seq2Seq models are less accurate in predicting \emph{all the slots} in a sentence (which is on average 6), but on the other hand, they excel in predicting individual slots, specially the type-II slots.
One interesting factor is, the Seq2Seq models significantly outperform sequence tagging models in predicting type-II slots.
Note that type-II slots are longer and less frequent, and we suspect conditional text generation helps Seq2Seq models predict them accurately.
In comparison, we suspect that due to fewer labeled examples of type-II slots, the sequence tagging models perform poorly on that category (as noted before, we train the sequence tagging models for the type-I and type-II slots individually).
% Later, we will show what types of mistakes these two modeling approaches make in prediction.

Next, we break down RoBERTa (w/ CRF) and BART's performances, the best performing models in their respective model categories, followed by an error analysis to shed light on the error types.

\input{tables/intent_analysis}

\subsection{Performance Breakdown} 
\label{sec:perf_break}

\paragraph{Intent Classification}
% \smallskip
% \noindent\textbf{Intent Classification\hspace{1.5ex}}
In the {\dataset} corpus, 38\% of the sentences fall into the first four categories: Data Collection, Data Sharing, Data Storage, Data Security, and the remaining belong to the Other category.
% We breakdown the test performances of the best joint intent and slot tagging model (RoBERTa) and Seq2Seq model (BART) to provide a thorough ablation analysis. We examine the model performances across different intent types, slot types and slot lengths.
Therefore, we investigate how much the models are confused in predicting the accurate intent label.
We provide the confusion matrix of the models in Appendix.
Due to an imbalanced distribution of labels, BART makes many incorrect predictions.
% Figure~\ref{fig:cm-bart} shows the confusion matrix of the BART model (see Appendix for the RoBERTa model), and we see, due to the imbalance distribution of labels, BART makes many incorrect predictions.
% for sentences with Data Sharing and Data Security labels.
We notice that BART is confused most between \emph{Data Collection} and \emph{Data Storage} labels. 
Our manual analysis reveals that BART is confused between slot labels \{``Data Collector'', ``Data Holder''\} and \{``Data Retained'', ``Data Collected''\} as they are often associated with the same text span.
We suspect this leads to BART's confusion.
Table~\ref{table:intent} presents the performance breakdown across intent labels. 

% For examples with intents such as Data Collection/Usage, Data Sharing/Disclosure and Data Security/Protection, the model most likely to miss classify them as ``Other''. In addition, those mis-classified examples with intent Data Storage/Retention tend to be predicted as Data Collection/Usage. Similar result appears in the confusion matrix of the BART model and we defer it to the appendix due to the space limit.

% \paragraph{Slot Filling}
%  Key points of Figure~\ref{figure:fig:cm-roberta}
% (1) ``Collection'', ``sharing'' and ``Security'' are mis-classified into ``Other''
% (2) ``retention'' is mis-classified into ``collection''
% In Table~\ref{table:intent}, we present the test performance breakdowns of RoBERTa and BART across different intent types. 
% RoBERTa and BART are comparable in terms of intent F1. However, BART outperforms RoBERTa by a large margin in terms of slot F1. We speculate the reason is that Seq2Seq learning models the problem as conditional text generation combining the contextual information from the input sequence and the decoded output sequence, while sequence tagging approach uses separate classifiers to predict intent and slot label for each input token. In this regard, Seq2Seq learning preserves more contextual information and thus gives better predictions.

% Key points of Table~\ref{table:intent}
% (1) Intent F1 comparable to each other
% (2) Slot F1: BART > RoBERTa
% (3) Some types of slots are harder to predict, e.g., type-II slot in data security/protection, type-I slots in data storage/retention.

\paragraph{Slot Filling}
% \smallskip
% \noindent\textbf{Slot Filling\hspace{1.5ex}}
We breakdown the models' performances in slot filling under two settings. 
First, Table~\ref{table:intent} shows slot filling performance under different intent categories.
Among the four classes, the models perform worst on slots associated with the ``Data Security'' intent class as {\dataset} has the lowest amount of annotations for that intent category.
Second, we demonstrate the models' performances on different slot types in Figure \ref{fig:slot_type_acc}.
RoBERTa's recall score for ``polarity'', ``protect-against'', ``protection-method'' and ``storage-place'' slot types is zero.
This is because these slot types have the lowest amount of training examples in {\dataset}.
On the other hand, BART achieves a higher recall score, specially for the ``polarity'' label as their corresponding spans are short.

We also study the models' performances on slots of different lengths.
The results show that BART outperforms RoBERTa by a larger margin on longer slots 
(see Figure \ref{fig:slot_lenth_acc}), 
% (see Figure \ref{fig:slot_lenth_acc} in Appendix), 
corroborating our hypothesis that conditional text generation results in more accurate predictions for longer spans.

\input{figures/slot_breakdown}
\input{figures/slot_breakdown_lengthwise}

\input{tables/error_analysis}
\input{tables/error_types}

\subsection{Error Analysis}
\label{sec:error_anal}

We analyze the incorrect intent and slot predictions by RoBERTa and BART.
We categorize the errors into seven types.
Note that a predicted slot is considered correct if its' label and span both match (\emph{exact match}) one of the references.
We characterize the error types as follows.

\begin{enumerate}
\setlength{\itemindent}{0em}
\item \textbf{Wrong Intent (WI)}: The predicted intent label does not match the reference intent label.
\item \textbf{Missing Slot (MS)}: None of the predicted slots \emph{exactly} match a reference slot.
\item \textbf{Spurious Slot (SS)}: Label of a predicted slot does not match any of the references.
\item \textbf{Wrong Split (WSp)}: Two or more predicted slot spans with the same label could be merged to match one of the reference slots. A merged span and a reference span may \emph{only} differ in punctuations or stopwords (e.g., and).
\item \textbf{Wrong Boundary (WB)}: A predicted slot span is a sub-string of the reference span or vice versa. The slot label must exactly match.
\item \textbf{Wrong Label (WL)}: A predicted slot span matches a reference, but the label does not.
\item \textbf{Wrong Slot (WS)}: All other types of errors fall into this category.
\end{enumerate}

We provide one example of each error type in Table~\ref{table:error_analysis}.
In Table \ref{table:error_types}, we present the counts for each error type made by RoBERTa and BART models.
The two most frequent error types are SS and MS.
While BART makes more SS errors, RoBERTa suffers from MS errors.
While both the models are similar in terms of total errors, BART makes more correct predictions resulting in a higher Recall score, as discussed before.
One possible way to reduce SS errors is by penalizing more on wrong slot label prediction than slot span.
On the other hand, reducing MS errors is more challenging as many missing slots have fewer annotations than others.
We provide more qualitative examples in Appendix (see Table \ref{table:err-analysis-typeI} and \ref{table:err-analysis-typeII}) .
% \footnote{We provide more qualitative examples in Appendix.}
% (see Table \ref{table:err-analysis-typeI} and \ref{table:err-analysis-typeII}).

In the error analysis, we exclude the test examples (sentences) with the intent label ``Other'' and no slots. 
Out of 1,041 test instances in {\dataset}, there are 682 instances with the intent label ``Other''. 
We analyze RoBERTa and BART's predictions on those examples separately to check if the models predict slots as we consider them as spurious slots.
While RoBERTa meets our expectation of performing highly accurate (correct prediction for 621 out of 682), BART also correctly predicts 594 out of 682 by precisely generating ``[IN:Other]''.
Overall the error analysis aligns with our anticipation that the Seq2Seq modeling technique has promise and should be further explored in future works. 

%% file: tables/intent_analysis.tex
% \begin{table}[ht]
% \centering
% \resizebox{\linewidth}{!}{%
% \begin{tabular}{l|c|c@{\hskip 0.01in} c}
% \hline
% \multirow{2}{*}{Intent labels} & \multirow{2}{*}{Intent F1} & \multicolumn{2}{c}{Slot F1}\\
% \cline{3-4}
%  & & \multicolumn{1}{c}{Entity} & \multicolumn{1}{c}{Complex} \\ 
% \hline
% { Data } & \multirow{2}{*}{ 74.14$_{\pm 1.08}$} &  \multirow{2}{*}{ 59.77$_{\pm 0.82}$} &
% \multirow{2}{*}{ 28.91$_{\pm 2.69}$} \\
% { Collection/Usage} &  & &\\
% { Data} & \multirow{2}{*}{ 67.16$_{\pm 2.03}$} &  \multirow{2}{*}{ 53.58$_{\pm 5.66}$} &
% \multirow{2}{*}{ 34.44$_{\pm 3.37}$} \\
% { Sharing/Disclosure} &  & &\\
% { Data} & \multirow{2}{*}{ 61.68$_{\pm 3.60}$} &  \multirow{2}{*}{ 40.08$_{\pm 3.66}$} &
% \multirow{2}{*}{ 31.56$_{\pm 3.11}$} \\
% { Storage/Retention} &  & &\\
% { Data} & \multirow{2}{*}{ 68.91$_{\pm 2.89}$} &  \multirow{2}{*}{ 53.90$_{\pm 4.92}$} &
% \multirow{2}{*}{ 21.92$_{\pm 2.54}$} \\
% { Security/Protection} &  & &\\
% \hline
% \end{tabular}
% }
% \caption{
% Performance of the RoBERTa baseline for each intent on the PolicyIE dataset (on test set). We average Intent F1 values over two joint models.
% % We trained all the models five times with different seeds and reporting average performances.
% }
% \label{table:intent}
% \end{table}

\begin{table}[t]
\centering
\resizebox{\linewidth}{!}{%
% \small
\setlength\tabcolsep{4pt} % default value: 6pt
\begin{tabular}{l|c|c c}
\hline
\multirow{2}{*}{Intent labels} & \multirow{2}{*}{Intent F1} & \multicolumn{2}{c}{Slot F1}\\
\cline{3-4}
 & & \multicolumn{1}{c}{Type-I} & \multicolumn{1}{c}{Type-II} \\ 
\hline
\multicolumn{4}{l}{RoBERTa} \\
\hline
Data Collection & {74.1$_{\pm 1.1}$} &  59.8$_{\pm 0.8}$ &
28.9$_{\pm 2.7}$ \\
Data Sharing & 67.2$_{\pm 2.0}$ & 53.6$_{\pm 5.7}$ &
34.4$_{\pm 3.4}$ \\
Data Storage & 61.7$_{\pm 3.6}$ &  40.1$_{\pm 3.7}$ &
31.6$_{\pm 3.1}$ \\
Data Security & {68.9$_{\pm 2.9}$} &  53.9$_{\pm 4.9}$ &
21.9$_{\pm 2.5}$ \\
\hline
\multicolumn{4}{l}{BART} \\
\hline
Data Collection & 73.5$_{\pm 2.3}$ & {67.0$_{\pm 4.2}$} & {56.2$_{\pm 2.8}$} \\
Data Sharing & {70.4$_{\pm 2.7}$} & {61.2$_{\pm 1.6}$} & {53.5$_{\pm 3.4}$}  \\
Data Storage & {63.1$_{\pm 4.7}$} & {56.2$_{\pm 8.2}$} & {64.9$_{\pm 2.5}$} \\
Data Security & 67.2$_{\pm 3.9}$ & {66.0$_{\pm 2.2}$} & {32.8$_{\pm 1.3}$} \\
\hline
\end{tabular}
}
\vspace{-1mm}
\caption{
Test performance of the RoBERTa and BART model for each intent type.
% We average the intent F1 scores over examples with type-I and type-II slots for the RoBERTa model.
% For the RoBERTa model, we average the intent F1 scores over examples with type-I and type-II slots.
% For brevity, we truncated the intent labels.
% We trained all the models five times with different seeds and reporting average performances.
}
\vspace{-2mm}
\label{table:intent}
\end{table}

%% file: figures/slot_breakdown.tex
\begin{figure}[t]
\captionsetup[subfigure]{labelformat=empty}
\centering
\vspace{-2mm}
% \hspace{-4mm}
\includegraphics[width=0.48\textwidth]{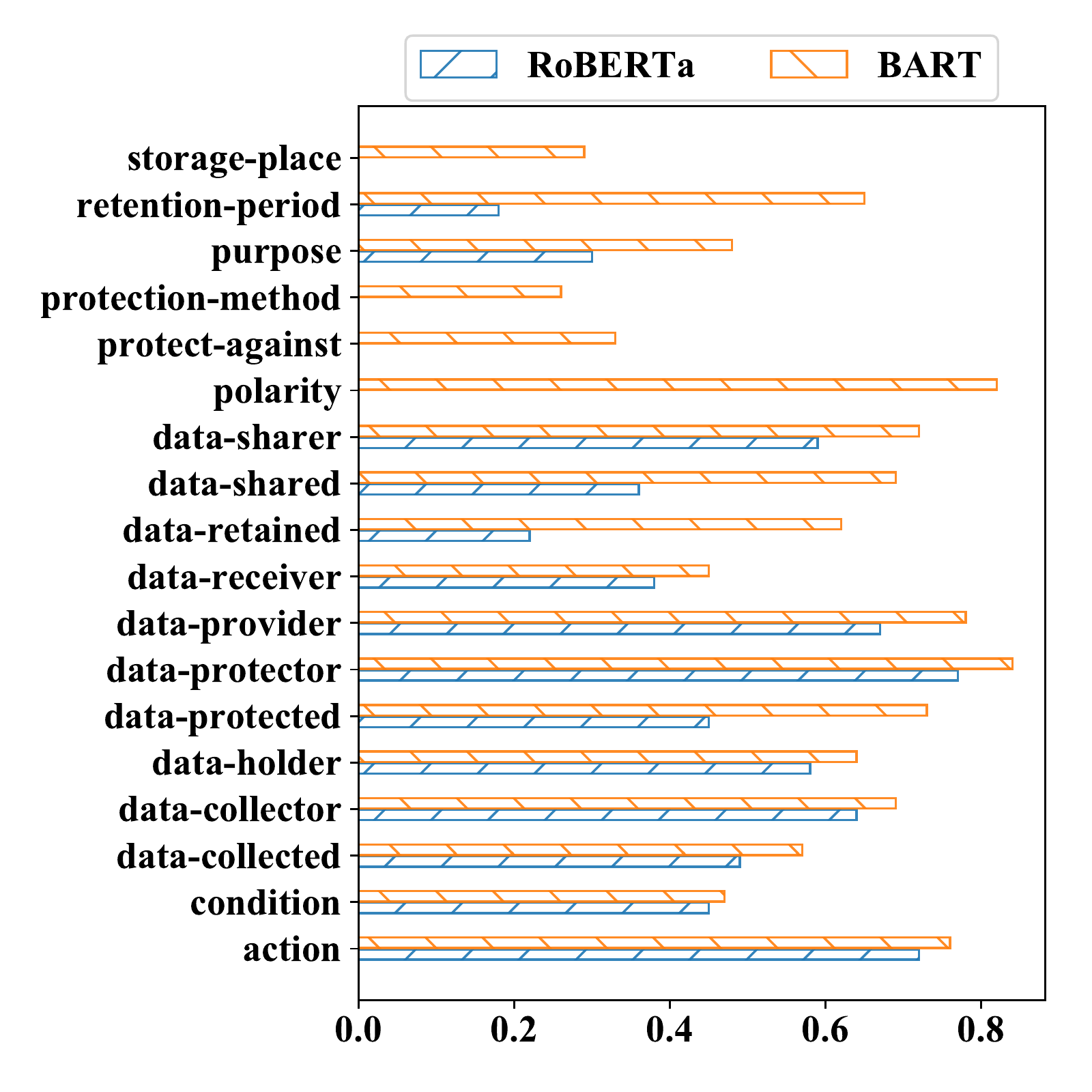}
\vspace{-7mm}
\caption{
% Comparison between best performing sequence tagging model (RoBERTa) and Seq2Seq model (BART) in terms of EM for different slot types.
Test set performance (Recall score) on {\dataset} for the eighteen slot types.
}
\vspace{-3mm}
\label{fig:slot_type_acc}
\end{figure}

% \begin{figure}[ht]
% \captionsetup[subfigure]{labelformat=empty}
% \centering
% \hspace{-5mm}
% \includegraphics[width=0.50\textwidth]{images/slot_accuracy_length_wise.pdf}
% \vspace{-3mm}
% \caption{
% Comparison between best performing sequence tagging model (RoBERTa) and Seq2Seq model (BART) in terms of EM for different slot lengths.
% }
% \label{fig:slot_lenth_acc}
% \end{figure}

%% file: figures/slot_breakdown_lengthwise.tex
\begin{figure}[ht]
\captionsetup[subfigure]{labelformat=empty}
\centering
% \hspace{-4mm}
\includegraphics[width=0.48\textwidth]{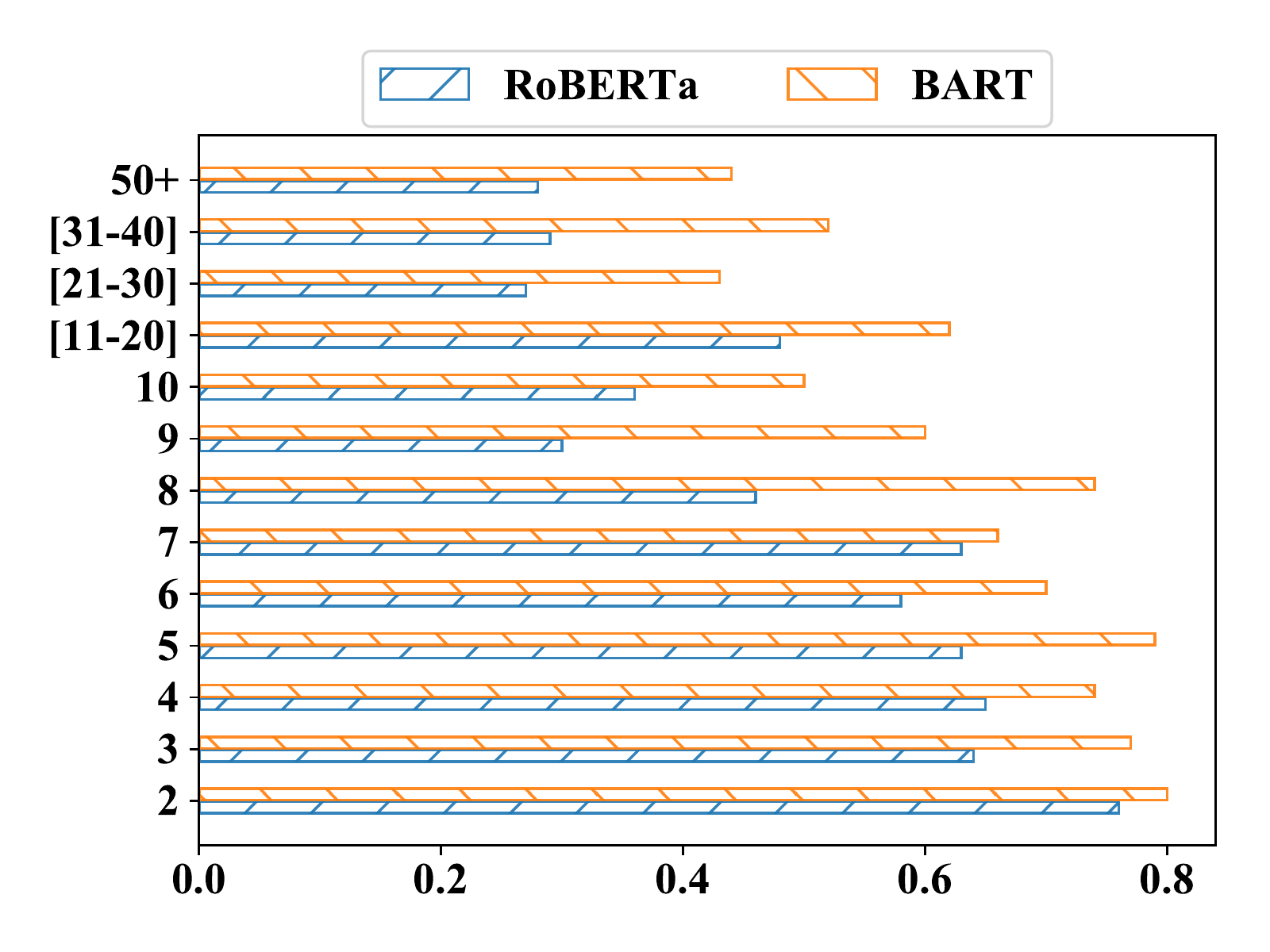}
\vspace{-7mm}
\caption{
% Comparison between best performing sequence tagging model (RoBERTa) and Seq2Seq model (BART) in terms of Recall for different slot lengths.
Test set performance (Recall score) on {\dataset} for slots with different length.
}
\vspace{-2mm}
\label{fig:slot_lenth_acc}
\end{figure}

%% file: tables/error_analysis.tex
\begin{table*}[!ht]
\centering
\resizebox{\linewidth}{!}{%
% \small
\def\arraystretch{1.2}%
\begin{tabular}{p{1.0\linewidth}}
\hline
% \textbf{Input:} However , please note that the third parties collect your information – and share to us , where applicable - based on their own privacy policies and applicable laws , which may have provisions and commitments different from this Privacy Policy .
% \\ \hdashline
$+$ {[IN:data-collection-usage [SL:data-provider.third-party-entity \emph{third parties}] [SL:action collect] [SL:data-provider.user \emph{your}] [SL:data-collected.data-general \emph{information}] [SL:data-collector.first-party-entity \emph{us}]]} \\
$-$ {[\textcolor{red}{IN:data-sharing-disclosure} [\textcolor{blue}{SL:data-receiver.third-party-entity} \emph{third parties}] [SL:action \textcolor{orange}{\emph{share}}] [SL:data-provider.user \emph{your}] [\textcolor{blue}{SL:data-shared.data-general} \emph{information}] [\textcolor{blue}{SL:data-sharer.first-party-entity} \emph{us}] [\textcolor{green}{SL:condition} \emph{where applicable}] [\textcolor{green}{SL:condition} \emph{based on their own privacy policies}]]} \\
\textbf{Error types}: \textcolor{red}{Wrong Intent (WI)}, \textcolor{blue}{Wrong Label (WL)}, \textcolor{orange}{Wrong Slot (WS)}, \textcolor{green}{Spurious Slot (SS)} \\
\hline
$+$ {[\ldots [\textcolor{chocolate}{SL:data-provider.third-party-entity \emph{third parties}}] [SL:condition \emph{it is allowed by applicable law or according to your agreement with third parties}]]} \\
$-$ {[\ldots [SL:condition \textcolor{magenta}{\emph{allowed by applicable law or according to your agreement with third parties}}]]} \\
\textbf{Error types}: \textcolor{magenta}{Wrong Boundary (WB)}, \textcolor{chocolate}{Missing Slot (MS)}  \\
% \hline
% $+$ {[\ldots [SL:action \emph{use}] [SL:data-provider.user \emph{your}] [SL:data-collected.data-general \emph{personal data}] [SL:purpose.legal-requirement \emph{establish a legal ground}]]} \\
% $-$ {[\ldots [SL:action \emph{use}] [SL:data-provider.user \emph{your}] [SL:data-collected.data-general \emph{personal data}] [SL:purpose.legal-requirement \textcolor{red}{\emph{establish a legal ground to use your personal data}}]]} \\
% \textbf{Error types}: \textcolor{red}{Bad Join (BJ)}  \\
\hline
% $+$ {[\ldots [SL:purpose.analytics-research \emph{improve your user experience and the overall quality of our services}]]} \\
% $-$ {[\ldots [SL:purpose.analytics-research \textcolor{purple}{\emph{improve your user experience}}] [SL:purpose.analytics-research \textcolor{purple}{\emph{the overall quality of our services}}]]} \\
% \textbf{Error types}: \textcolor{purple}{Wrong Split (WSp)}  \\
% \hline
$+$ {[\ldots [SL:data-receiver.third-party-entity \emph{social media and other similar platforms}] \ldots]} \\
$-$ {[\ldots [SL:data-receiver.third-party-entity \textcolor{purple}{\emph{social media}}] [SL:data-receiver.third-party-entity \textcolor{purple}{\emph{other similar platforms}}] \ldots]} \\
\textbf{Error types}: \textcolor{purple}{Wrong Split (WSp)}  \\
\hline
% $+$ {[\ldots [SL:purpose-argument.analytics-research \emph{measure traffic}] [SL:purpose-argument.analytics-research \emph{usage trends for the Service}]]} \\
% $-$ {[\ldots [SL:purpose-argument.analytics-research \textcolor{red}{\emph{measure traffic and usage trends for the Service}}]]} \\
% \textbf{Error types}: \textcolor{red}{Wrong Join (WJ)}  \\
% \hline
\end{tabular}
}
\vspace{-1mm}
\caption{
Three examples showing different error types appeared in BART's predictions. 
$+$ and $-$ indicates the reference and predicted sequences, respectively.
Best viewed in color.
}
% \vspace{-2mm}
\label{table:error_analysis}
\end{table*}

%% file: tables/error_types.tex
\begin{table}[!ht]
\centering
% \resizebox{\linewidth}{!}{%
% \small
% \def\arraystretch{1.1}%
\begin{tabular}{l|c|c}
\hline
Error & RoBERTa & BART \\
\hline
Wrong Intent & 161 & 178 \\
Spurious Slot & 472 & 723 \\
Missing Slot & 867 & 517 \\
Wrong Boundary & 130 & 160 \\
Wrong Slot & 103 & 143 \\
Wrong Split & 32 & 27 \\ 
Wrong Label & 18 & 19 \\ \hline
Total Slots & 2,198 & 2,198 \\
Correct Prediction & 1,064 & 1,361 \\
Total Errors & 1,622 & 1,589 \\
Total Predictions & 2,686 & 2,950 \\
\hline
\end{tabular}
% }
\vspace{-1mm}
\caption{
Counts for each error type on the test set of {\dataset} using RoBERTa and BART models. 
}
% }
\label{table:error_types}
\vspace{-2mm}
\end{table}

% \begin{table}[!ht]
% \centering
% % \resizebox{\linewidth}{!}{%
% % \small
% % \def\arraystretch{1.1}%
% \begin{tabular}{l|c|c}
% \hline
% Error & RoBERTa & BART \\
% \hline
% WI & \myprogbar{0.75} & \myprogbar{0.08} \\
% WSL & \myprogbar{0.75} & \myprogbar{0.01} \\
% WS & \myprogbar{0.75} & \myprogbar{0.07} \\
% SS & \myprogbar{0.75} & \myprogbar{0.59} \\
% WB & \myprogbar{0.75} & \myprogbar{0.08} \\
% MS & \myprogbar{0.75} & \myprogbar{0.25} \\
% WSp & \myprogbar{0.75} & \myprogbar{0.01} \\ \hline
% Total Errors & & 2098 \\
% \hline
% \end{tabular}
% % }
% \caption{
% Counts for each error type on the test set of \dataset. Bars indicate the number of errors, with white as zero and fully filled as the number in the Total Errors row.
% }
% % }
% \label{table:error_types_copy}
% % \vspace{-2mm}
% \end{table}

%% file: sections/relwork.tex
\section{Related Work}

% \smallskip
% \noindent\textbf{Automated Privacy Policy Analysis\hspace{1.5ex}}
\paragraph{Automated Privacy Policy Analysis}
Automating privacy policy analysis has drawn researchers' attention as it enables the users to know their rights and act accordingly. 
Therefore, significant research efforts have been devoted to understanding privacy policies.
Earlier approaches \cite{costante2012websites} designed rule-based pattern matching techniques to extract specific types of information.
% from privacy policies.
Under the \emph{Usable Privacy Project}~\citep{sadeh2013usable}, several works have been done~\citep{bhatia2015towards, wilson2016creation, wilson2016crowdsourcing, sathyendra2016automatic, bhatia2016automated, hosseini2016lexical,sathyendra2017identifying, zimmeck2019maps, bannihatti2020finding}.
Notable works leveraging NLP techniques include text alignment~\citep{liu2014step, ramanath2014unsupervised}, text classification~\citep{wilson2016creation, harkous2018polisis, zimmeck2019maps}, and question answering (QA)~\citep{shvartzshanider2018recipe, harkous2018polisis, Ravichander2019Question, ahmad-etal-2020-policyqa}. 
% and named entity recognition \cite{bokaie-hosseini-etal-2020-identifying}. 
% Specifically, \citet{costante2012websites} propose rule-based pattern matching techniques to extract name entity related to data collection;
% \citet{bokaie-hosseini-etal-2020-identifying} use named entity recognition models to extract third party entity mentioned in the documents. 
\citet{bokaie-hosseini-etal-2020-identifying} is the most closest to our work that used named entity recognition (NER) modeling technique to extract third party entities mentioned in policy documents.

% \paragraph{Benchmark Corpus}
% In this work, we propose {\dataset} corpus to enable information extraction from privacy policies by formulating the task as identifying the privacy practice behind every sentence in a policy document and predicting the constituent text spans that provide specific information.
Our proposed {\dataset} corpus is distinct from the previous privacy policies benchmarks: OPP-115~\citep{wilson2016creation} uses a hierarchical annotation scheme to annotate text segments with a set of data practices and it has been used for multi-label classification~\citep{wilson2016creation, harkous2018polisis} and question answering~\citep{harkous2018polisis, ahmad-etal-2020-policyqa}; PrivacyQA~\citep{Ravichander2019Question} frame the QA task as identifying a list of relevant sentences from policy documents. 
Recently, \citet{bui2021automated} created a dataset by tagging documents from OPP-115 for privacy practices and uses NER models to extract them.
% Similarly, a concurrent work \cite{bui2021automated} uses NER models to extract privacy practices.
In contrast, {\dataset} is developed by following semantic parsing benchmarks, and we model the task following the NLP literature.

% to build dialogue systems.
% In contrast, {\dataset} is developed by following task-oriented semantic parsing benchmarks used in NLP literature to build dialogue systems.
% Compared to prior benchmarks for privacy policies, {\dataset} corpus is a task-oriented semantic parsing dataset which provides fine-grained and structured annotations.

% Difference between OPP-115 corpus and {\dataset~corpus}: OPP-115 corpus provides a hierarchical annotation scheme which annotate a text segment from privacy policies associated with a set of data practice labels and it is used for multi-label classification~\citep{wilson2016creation, harkous2018polisis} and question answering~\citep{harkous2018polisis, ahmad-etal-2020-policyqa}. 

% Difference between PrivacyQA and {\dataset~corpus}: PrivacyQA frame the QA task as extracting a list of sentences from policy documents given a question.  

\paragraph{Intent Classification and Slot Filling}
% \smallskip
% \noindent\textbf{Intent Classification and Slot Filling\hspace{1.5ex}}
Voice assistants and chat-bots frame the task of natural language understanding via classifying intents and filling slots given user utterances. 
% ATIS~\citep{hemphill1990atis}, SNIPS~\citep{coucke2018snips}, TOP~\cite{gupta-etal-2018-semantic} are among the popular semantic parsing benchmarks. Recent works proposed new benchmarks supporting more domains and languages~\citep{upadhyay2018almost, schuster2019cross, xu-etal-2020-end, li2020mtop}. 
Several benchmarks have been proposed in literature covering several domains, and languages \cite{hemphill1990atis, coucke2018snips, gupta-etal-2018-semantic, upadhyay2018almost, schuster2019cross, xu-etal-2020-end, li2020mtop}. 
Our proposed {\dataset} corpus is a new addition to the literature within the security and privacy domain. 
{\dataset} enables us to build conversational solutions that users can interact with and learn about privacy policies.
% To the best of our knowledge, intent classification and slot filling tasks has never been explored in the domain of privacy privacy documents. 
% Our work takes a first step towards understanding privacy policies by presenting {\dataset} corpus and identifying challenges in this domain.

% ATIS~\citep{hemphill1990atis} contains natural language commands of flight reservations. 
% ATIS label (e.g., cheapest, capacity, airline) and slot label (e.g., airline code, arrive_date) 
% \citet{coucke2018snips} SNIPS-NLU contains natural language commands collected by SNIPS personal voice assistant. 

% \citet{upadhyay2018almost} % it contains 4978 English utterances from the English ATIS Corpus for training and translated and annotated 600 examples in Turkish and Hindi respectively for supervision using human translators and Amazon Mechanical Turk

% \citet{schuster2019cross} % Multilingual TOP (Schuster et al., 2018) is a multilingual task-oriented parsing dataset of 57k annotated utterances in English (43k), Spanish (8.6k) and Thai (5k) across the domains weather, alarm, and reminder.

% \citet{li2020mtop} MTOP is a multilingual task-oriented semantic parsing dataset covering 6 languages and 11 domains. MTOP also allows compositional queries by introducing nested slots in their data representation.

%% file: sections/conclusion.tex
\section{Conclusion}
This work aims to stimulate research on automating information extraction from privacy policies and reconcile it with users' understanding of their rights.
We present \dataset, an intent classification and slot filling benchmark on privacy policies with two alternative neural approaches as baselines.
We perform a thorough error analysis to shed light on the limitations of the two baseline approaches. 
% Two alternative neural modeling approaches are set as baselines on \dataset~and a thorough error analysis is performed to shed light on the limitations. 
We hope this contribution would call for research efforts in the specialized privacy domain from both privacy and NLP communities.
% with 5,250 intent and 11,788 slot annotations.

%% file: appendix.tex
\input{tables/slot_subtypes}
\input{tables/privacy_practices}

\input{figures/confusion_matrix}
% \input{figures/slot_analysis}

% \section{Qualitative Examples}
\input{tables/error_analysis_typeI}

\input{tables/error_analysis_typeII}

% \end{document}

%% file: tables/slot_subtypes.tex
\begin{table*}[h]
\centering
\def\arraystretch{1.2}%
\resizebox{\linewidth}{!}{%
\begin{tabular}{l|l}
\hline
Type-I slots & Attributes\\
\hline
Action & None\\
\hline
Data Provider & (1) User (2) Third party entity \\
\hline
Data Collector & (1) First party entity\\
\hline
Data Collected & (1) General Data (2) Aggregated/Non-identifiable data (3) Contact data (4) Financial data \\
& (5) Location data (6) Demographic data (7) Cookies, web beacons and other technologies \\
& (8) Computer/Device data (9) User online activities/profiles  (10) Other data \\
\hline
Data Sharer & (1) First party entity \\
\hline 
Data Shared & (1) General Data (2) Aggregated/Non-identifiable data (3) Contact data (4) Financial data \\
& (5) Location data (6) Demographic data (7) Cookies, web beacons and other technologies \\
& (8) Computer/Device data (9) User online activities/profiles  (10) Other data \\
\hline
Data Receiver & (1) Third party entity\\
\hline
Data Holder & (1) First party entity (2) Third party entity \\
\hline
Data Retained & (1) General Data (2) Aggregated/Non-identifiable data (3) Contact data (4) Financial data \\
& (5) Location data (6) Demographic data (7) Cookies, web beacons and other technologies \\
& (8) Computer/Device data (9) User online activities/profiles  (10) Other data \\
\hline
Storage Place & None \\
\hline
Retention Period & None\\
\hline
Data Protector & (1) First party entity (2) Third party entity \\
\hline
Data Protected & (1) General Data (2) Aggregated/Non-identifiable data (3) Contact data (4) Financial data \\
& (5) Location data (6) Demographic data (7) Cookies, web beacons and other technologies \\
& (8) Computer/Device data (9) User online activities/profiles  (10) Other data \\
\hline
Protect Against & Security threat \\
\hline
\hline
Type-II slots & Attributes\\
\hline
Purpose & (1) Basic service/feature (2) Advertising/Marketing (3) Legal requirement\\
& (4) Service operation and security (5) Personalization/customization\\
& (6) Analytics/research (7) Communications (8 Merge/Acquisition (9) Other purpose \\
\hline
Condition & None \\
\hline
Polarity & (1) Negation \\
\hline
Protection Method & (1) General safeguard method (2) User authentication (3) Access limitation\\
&(5) Encryptions (6) Other protection method \\
\hline
\end{tabular}
}
% \vspace{-2mm}
\caption{Slots and their associated attributes. ``None'' indicates there are no attributes for the those slots.}
\label{table:slot-attribute}
\end{table*}

%% file: tables/privacy_practices.tex
%%%%%%%%%%%%%%%%%%%%%%%%%%%%%%%%%%%%%%%%%%%%%%%%%%%%%%%%%%%%%%%%
% \begin{table*}[!htb]
% \centering
% \caption{Privacy Practices and the Corresponding Arguments. (split the arguments into two category)}
% \resizebox{\textwidth}{!}{
% \begin{tabular}{l|c c c c} 
% \hline
%  \begin{tabular}[x]{@{}c@{}}Privacy\\Practices\end{tabular} & \begin{tabular}[x]{@{}c@{}}Data\\Collection/Usage\end{tabular}  & \begin{tabular}[x]{@{}c@{}}Data\\Sharing/Disclosure\end{tabular} & \begin{tabular}[x]{@{}c@{}}Data\\Storage/Retention/Deletion\end{tabular} & \begin{tabular}[x]{@{}c@{}}Data\\Security/Protection\end{tabular} \\ 
% \hline
% \hline
% Arg. 1 & Purpose & Purpose & Purpose  & Purpose \\ 
% % \hline
% Arg. 2 & Condition & Condition & Condition & Condition \\ 
% % \hline
% Arg. 3 & Polarity & Polarity & Polarity & Polarity \\ 
% Arg. 4 & Action & Action & Action & Action \\ 
% % \hline
% Arg. 5 & Data Collector & Data Sharer & Data Holder & Data Protector \\ 
% % \hline
% Arg. 6 & Data Provider & Data Provider & Data Provider & Data Provider \\ 
% % \hline
% Arg. 7 & Data Collected & Data Shared & Data Retained & Data Protected \\ 
% % \hline
% Arg. 8 & N/A & Data Receiver & Storage Place & Protection Method \\ 
% % \hline
% Arg. 9 & N/A & N/A & Retention Period & Protect Against \\
% % \hline

% \hline
% \end{tabular}
% }
% \label{tab:privacy-practice}
% \end{table*}

\begin{table*}[h]
\centering
\resizebox{\textwidth}{!}{
\begin{tabular}{l |c c c c} 
\multirow{2}{*}{Privacy Practices} & Data & Data & Data & Data \\
& Collection/Usage & Sharing/Disclosure & Storage/Retention & Security/Protection \\
\hline
\multicolumn{5}{l}{Type-I slots} \\
\hline
Action & 750 / 169 & 344 / 70 & 198 / 57 & 102 / 31 \\
Data Provider & 784 / 172 & 247 / 54 & 139 / 44 & 65 / 20 \\
% \hline
Data Collector & 653 / 151 & - & - & - \\
Data Collected & 1833 / 361 & - & - & - \\
% \hline
Data Sharer & - & 288 / 54 & - & - \\
Data Shared & - & 541 / 110 & - & - \\
Data Receiver & - & 456 / 115 & - & - \\
% \hline
Data Holder & - & - & 192 / 59 & - \\
Data Retained & - & - & 291 / 119 & - \\
Storage Place & - & - & 70 / 21 & - \\
Retention Period & - & - & 101 / 17 & - \\
% \hline
Data Protector & - & - & - & 105 / 31 \\
Data Protected & - & - & - & 119 / 34 \\
Protect Against & - & - & - & 49 / 15 \\
\hline
\multicolumn{5}{l}{Type-II slots} \\
\hline
Purpose & 894 / 193 & 327 / 65 & 168 / 40 & 5 / 0 \\
Condition & 337 / 81 & 154 / 26 & 81 / 25 & 43 / 7 \\
Polarity & 50 / 15 & 21 / 1 & 22 / 1 & 18 / 5 \\
Protection Method & - & - & - & 143 / 35 \\
\hline
\# of slots & 5301 / 1142 & 2378 / 495 & 1262 / 383 & 649 / 178 \\
\hline
\# of sequences & 919 / 186 & 380 / 83 & 232 / 61 & 103 / 29 \\
\hline
\end{tabular}
}
% \vspace{-2mm}
\caption{Privacy practices and the associated slots with their distributions. ``X / Y'' indicates there are X instances in the train set and Y instances in the test set.}
% \vspace{-2mm}
\label{table:privacy-practice}
\end{table*}

%% file: figures/confusion_matrix.tex
\begin{figure*}[!ht]
\centering
\begin{minipage}{.48\textwidth}
  \centering
  \captionsetup{justification=centering}
  \includegraphics[width=\linewidth]{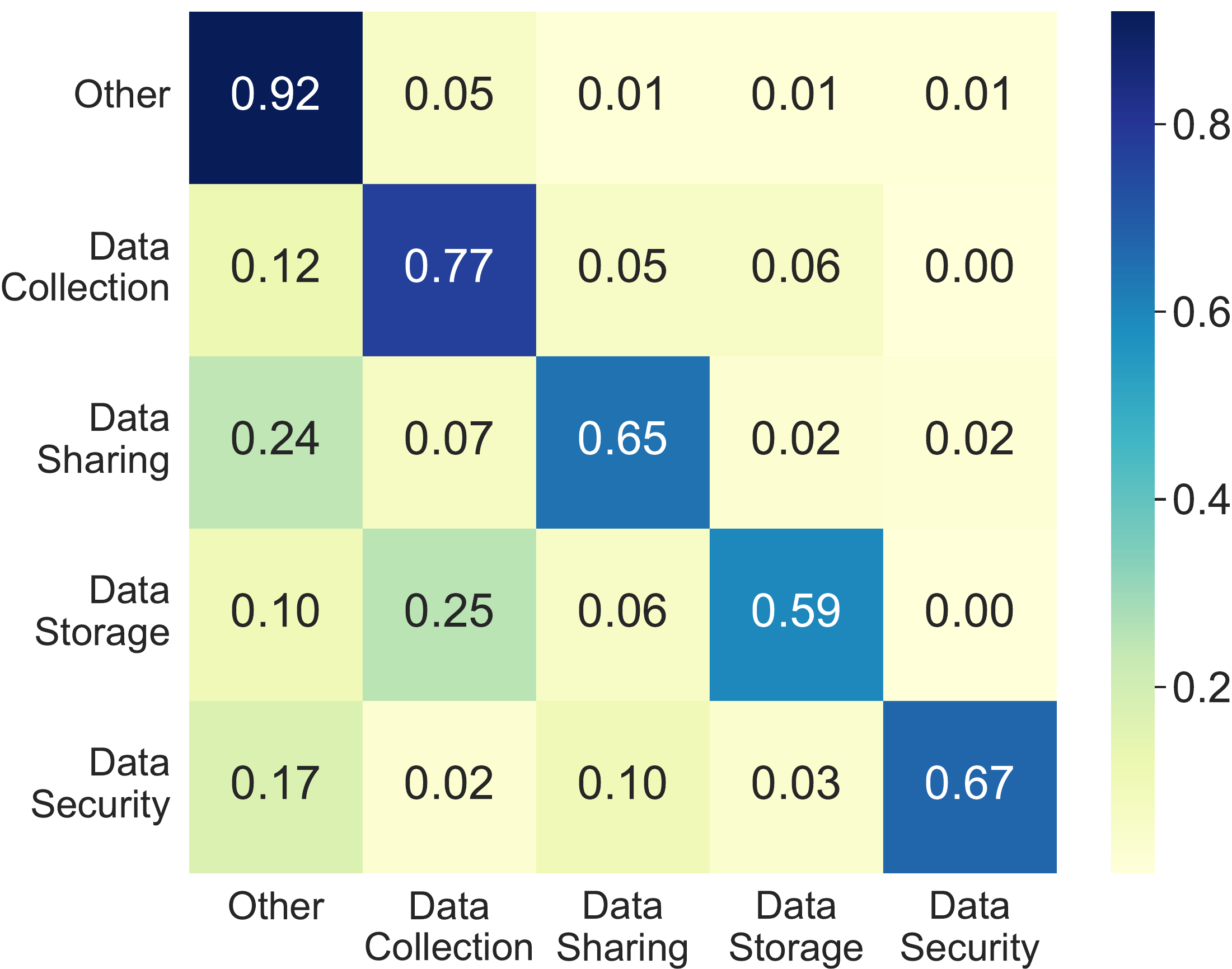}
  \captionof{figure}{Confusion matrix for intent classification using the RoBERTa model.}
  \label{fig:cm-roberta}
\end{minipage}%
\hspace{1mm}
\begin{minipage}{.48\textwidth}
  \centering
  \captionsetup{justification=centering}
  \includegraphics[width=\linewidth]{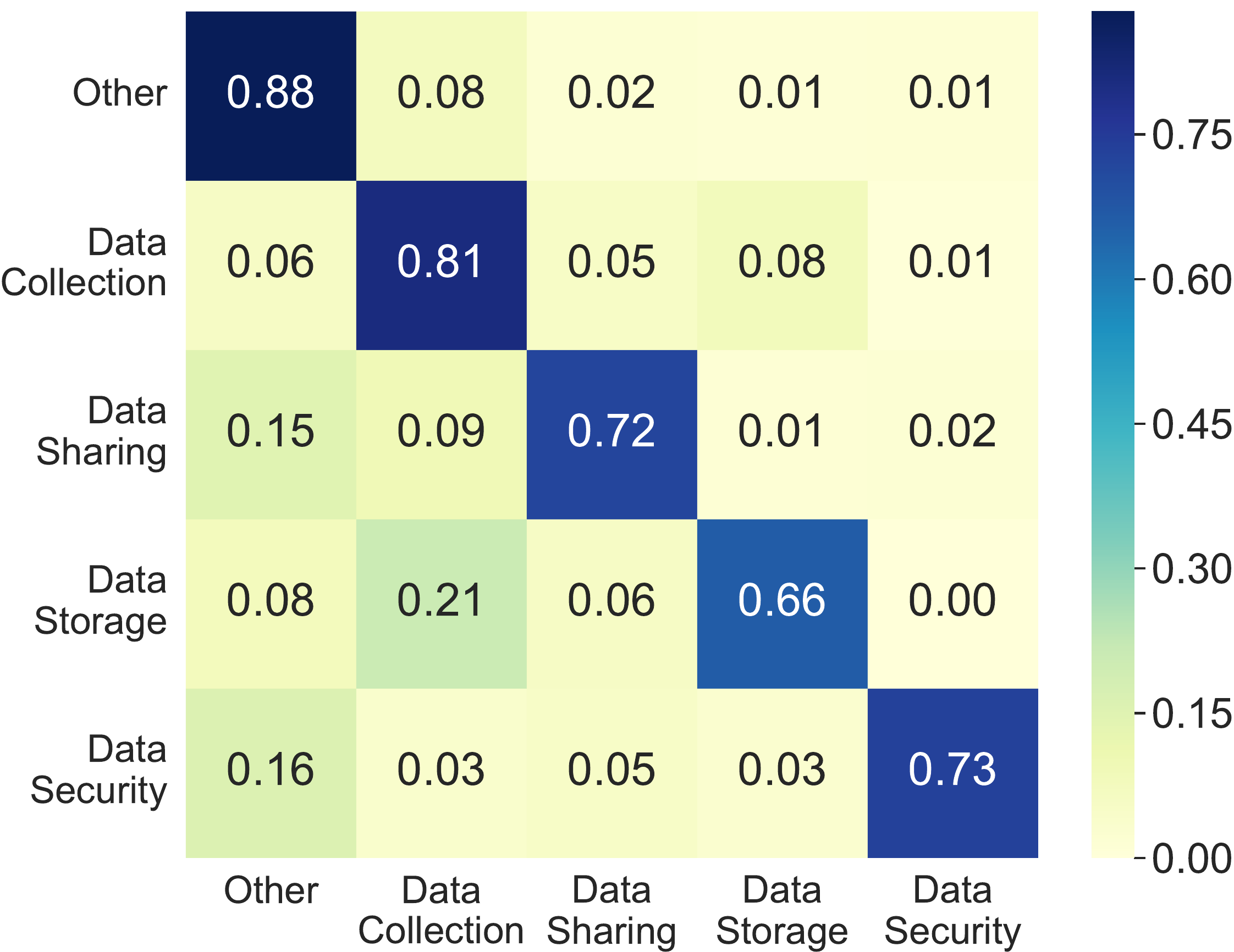}
  \captionof{figure}{Confusion matrix for intent classification using the BART model.}
  \label{fig:cm-bart}
\end{minipage}
\end{figure*}

%% file: tables/error_analysis_typeI.tex
\begin{table*}
\centering
\resizebox{\linewidth}{!}{%
\begin{tabular}{c|c|p{0.4\linewidth}|p{0.4\linewidth}}
& & Label & Text \\
\hline
\multirow{4}{*}{Ground truth} & & data-holder.first-party-entity & We \\ \cdashline{2-4}
& & action & keep \\ \cdashline{2-4} 
& &  data-retained.data-general & records \\ \cdashline{2-4} 
& & retention-period.retention-period & a period of no more than 6 years \\ 
\hline
\multirow{3}{*}{\makecell{RoBERTa \\ (P:1.0, R: 0.75)}} & \cmark &  data-holder.first-party-entity & We \\ \cdashline{2-4}
& \cmark &  action &  keep \\ \cdashline{2-4} 
& \cmark &  retention-period.retention-period &  a period of no more than 6 years \\ 
\hline
\multirow{3}{*}{\makecell{BART \\ (P:1.0, R: 1.0)}} & \cmark &  data-holder.first-party-entity &  We \\ \cdashline{2-4}
& \cmark &  action &  keep \\ \cdashline{2-4} 
& \cmark &  data-retained.data-general &  records \\ \cdashline{2-4} 
& \cmark &  retention-period.retention-period &  a period of no more than 6 years \\ 
\hline
\hline
\multirow{3}{*}{Ground truth} & &  data-collector.first-party-entity &  We \\ \cdashline{2-4}
& &  action &  access \\ \cdashline{2-4} 
& &  data-collected.data-general &  information \\ 
\hline
\multirow{2}{*}{\makecell{RoBERTa \\ (P:0.0, R: 0.0)}} & \xmark &  data-sharer.first-party-entity &  We \\ \cdashline{2-4}
& \xmark &  data-shared.data-general  &  information \\ 
\hline
\multirow{3}{*}{\makecell{BART \\ (P:0.0, R: 0.0)}} & \xmark &  data-sharer.first-party-entity &  We \\ \cdashline{2-4}
& \xmark &  action &  disclose \\ \cdashline{2-4} 
& \xmark &  data-shared.data-general &  information \\ 
\hline
\hline
\multirow{5}{*}{Ground truth} & &  data-sharer.first-party-entity &  Marco Polo \\ \cdashline{2-4}
& &  data-receiver.third-party-entity &  third party \\ \cdashline{2-4} 
& &  data-shared.data-general &  Personal Information \\ \cdashline{2-4} 
& &  data-provider.user &  users \\ \cdashline{2-4} 
& &  action &  transferred \\ 
\hline
\multirow{5}{*}{\makecell{RoBERTa \\ (P:0.6, R: 0.6)}} & \xmark &  data-receiver.third-party-entity &  Marco \\ \cdashline{2-4}
& \xmark &  data-sharer.first-party-entity  &  our \\ \cdashline{2-4}
& \cmark &  data-receiver.third-party-entity  &  third party \\ \cdashline{2-4}
& \cmark &  data-shared.data-general  &  Personal Information \\ \cdashline{2-4}
& \cmark &  action  &  transferred \\ 
\hline
\multirow{6}{*}{\makecell{BART \\ (P:0.83, R: 1.0)}} & \cmark &  data-sharer.first-party-entity &  Marco Polo \\ \cdashline{2-4}
& \cmark &  data-receiver.third-party-entity &  third party \\ \cdashline{2-4} 
& \cmark &  data-shared.data-general &  Personal Information \\ \cdashline{2-4} 
& \xmark &  data-sharer.first-party-entity &  us \\ \cdashline{2-4} 
& \cmark &  data-provider.user &  users \\ \cdashline{2-4} 
& \cmark &  action &  transferred \\ 
\hline
\hline 
% \multirow{4}{*}{Ground truth} & &  data-collector.first-party-entity &  We \\ \cdashline{2-4}
% & &  action &  retain \\ \cdashline{2-4} 
% & &  data-provider.user &  user \\ \cdashline{2-4} 
% & &  data-collected.data-contact &  contact information \\ 
% \hline
% \multirow{5}{*}{\makecell{RoBERTa \\ (P:0.25, R: 0.25)}} & \xmark &  data-holder.first-party-entity  &  We \\ \cdashline{2-4}
% & \cmark &  action  &  retain \\ \cdashline{2-4}
% & \xmark &  data-retained.data-general  &  basic \\ \cdashline{2-4}
% & \xmark &  data-retained.data-general  &  user contact information \\ \cdashline{2-4}
% & \xmark &  action  &  transferred \\ 
% \hline
% \multirow{3}{*}{\makecell{BART \\ (P:0.33, R: 0.25)}} & \xmark &  data-holder.first-party-entity &  We \\ \cdashline{2-4}
% & \cmark &  action & retain \\ \cdashline{2-4} 
% & \xmark &  data-retained.data-user-online-activities-profiles &  basic user contact information \\
% \hline
% \hline 
\multirow{4}{*}{Ground truth} & &  data-sharer.first-party-entity &  We \\ \cdashline{2-4}
& &  data-receiver.third-party-entity &  third parties \\ \cdashline{2-4} 
& &  action &  provide \\ \cdashline{2-4} 
& &  data-shared.data-general & information \\ 
\hline
\multirow{4}{*}{\makecell{RoBERTa \\ (P:1.0, R: 1.0)}} & \cmark &  data-sharer.first-party-entity  &  We \\ \cdashline{2-4}
& \cmark &  data-receiver.third-party-entity  &  third parties \\ \cdashline{2-4}
& \cmark &  action  &  provide \\ \cdashline{2-4}
& \cmark &  data-shared.data-general  &  information \\ 
\hline
\multirow{4}{*}{\makecell{BART \\ (P:0.25, R: 0.25)}} & \xmark &  data-collector.first-party-entity &  We \\ \cdashline{2-4}
& \xmark &  data-provider.third-party-entity & third parties \\ \cdashline{2-4}
& \cmark &  action & provide \\ \cdashline{2-4} 
& \xmark &  data-collected.data-general &  information \\
\hline
\end{tabular}
}
\caption{
Sample RoBERTa and BART predictions of Type-I slots.
(\cmark) and (\xmark) indicates correct and incorrect predictions, respectively.
Precision (P) and recall (R) score is reported for each example in the left column. 
}
\label{table:err-analysis-typeI}
\end{table*}

%% file: tables/error_analysis_typeII.tex
\begin{table*}
\centering
\resizebox{\linewidth}{!}{%
\begin{tabular}{c|c|p{0.9\linewidth}}
\hline
\multirow{2}{*}{Ground truth} & & [Label] condition \\
& & [Text] you use our product and service or view the content provided by us \\ \hline
\multirow{1}{*}{\makecell{RoBERTa \\ (P:1.0, R: 1.0)}} & \multirow{2}{*}{\cmark} & [Label] condition \\
& & [Text] you use our product and service or view the content provided by us  \\ \hline
\multirow{1}{*}{\makecell{BART \\ (P:1.0, R: 1.0)}} & \multirow{2}{*}{\cmark} & [Label] condition \\
& & [Text] you use our product and service or view the content provided by us \\
\hline
\hline
\multirow{4}{*}{Ground truth} & & [Label] purpose.other \\
& & [Text] their own purposes \\ \cdashline{2-3} 
& & [Label] purpose.advertising-marketing \\
& & [Text ] inform advertising related services provided to other clients \\ \hline
\multirow{2}{*}{\makecell{RoBERTa \\ (P:0.0, R: 0.0)}}  & \multirow{2}{*}{\xmark} & [Label] None  \\ 
& & [Text] None \\ \hline
\multirow{4}{*}{\makecell{BART \\ (P:1.0, R: 1.0)}} & \multirow{2}{*}{\cmark} & [Label] purpose.other \\
& & [Text] their own purposes \\ \cdashline{2-3} 
& \multirow{2}{*}{\cmark} & [Label] purpose.advertising-marketing \\
 & & [Text] inform advertising related services provided to other clients \\
\hline
\hline
\multirow{10}{*}{Ground truth} & & [Label] purpose.personalization-customization \\
& & [Text] provide more tailored services and user experiences \\ \cdashline{2-3}
& & [Label] purpose.basic-service-feature \\
& & [Text] remembering your account identity \\ \cdashline{2-3} 
& & [Label] purpose.service-operation-and-security \\
& & [Text] analyzing your account 's security \\ \cdashline{2-3} 
& & [Label] purpose.analytics-research \\
& & [Text] analyzing your usage of our product and service \\ \cdashline{2-3} 
& & [Label] purpose.advertising-marketing \\
& & [Text] advertisement optimization ( helping us to provide you with more targeted advertisements instead of general advertisements based on your information ) \\
\hline
\multirow{12}{*}{\makecell{RoBERTa \\ (P:0.17, R: 0.2)}} & \multirow{2}{*}{\xmark} & [Label] purpose.basic-service-feature \\
& & [Text] provide \\ \cdashline{2-3} 
& \multirow{2}{*}{\xmark} & [Label] purpose.other \\
& & [Text] purposes \\ \cdashline{2-3} 
& \multirow{2}{*}{\xmark} & [Label] purpose.analytics-research \\
& & [Text] remembering your account identity \\ \cdashline{2-3} 
& \multirow{2}{*}{\xmark} & [Label] purpose.analytics-research \\
& & [Text] analyzing your account ’s security \\ \cdashline{2-3} 
& \multirow{2}{*}{\cmark} & [Label] purpose.analytics-research \\
& & [Text] analyzing your usage of our product and service \\ \cdashline{2-3} 
& \multirow{2}{*}{\xmark} & [Label] purpose.advertising-marketing \\
& & [Text] advertisement optimization \\ 
\hline
\multirow{14}{*}{\makecell{BART \\ (P:0.43, R: 0.6)}} & \multirow{2}{*}{\cmark} &
[Label] purpose.personalization-customization \\
& & [Text] provide more tailored services and user experiences \\ \cdashline{2-3} 
& \multirow{2}{*}{\xmark} & [Label] purpose.service-operation-and-security \\
& & [Text] remembering your account identity \\ \cdashline{2-3} 
& \multirow{2}{*}{\cmark} & [Label] purpose.service-operation-and-security \\
& & [Text] analyzing your account ’s security \\ \cdashline{2-3} 
& \multirow{2}{*}{\cmark} & [Label] purpose.analytics-research \\
& & [Text] analyzing your usage of our product and service \\ \cdashline{2-3} 
& \multirow{2}{*}{\xmark} & [Label] purpose.advertising-marketing \\
& & [Text] advertisement optimization \\ \cdashline{2-3} 
& \multirow{2}{*}{\xmark} & [Label] purpose.advertising-marketing \\
& & [Text] provide you with more targeted advertisements instead of general advertisements \\ \cdashline{2-3} 
& \multirow{2}{*}{\xmark} & [Label] purpose.advertising-marketing \\
& & [Text] based on your information \\ 
\hline
\end{tabular}
}
\caption{
Sample RoBERTa and BART predictions of Type-II slots.
(\cmark) and (\xmark) indicates correct and incorrect predictions, respectively.
Precision (P) and recall (R) score is reported for each example in the left column. 
}
\label{table:err-analysis-typeII}
\end{table*}